\documentclass[lettersize,journal]{IEEEtran}
\usepackage{amsmath,amsfonts}
\usepackage{algorithmic}
\usepackage{algorithm}
\usepackage{array}
\usepackage[caption=false,font=normalsize,labelfont=sf,textfont=sf]{subfig}
\usepackage{textcomp}
\usepackage{stfloats}
\usepackage{url}
\usepackage{verbatim}
\usepackage{graphicx}
\usepackage{cite}
\usepackage{newtxtext,newtxmath}
\usepackage{float}
\usepackage{multirow}
\usepackage{booktabs}
\usepackage{wrapfig}

\begin{document}

\title{OmniDialog: An Omnipotent Pre-training \\ Model for Task-Oriented Dialogue System}

\author{Mingtao Yang, See-Kiong Ng, Jinlan Fu$^{*}$\thanks{\ \ * Corresponding author}

\thanks{Mingtao Yang, and See-Kiong Ng are with the Department of Computer Science, School of Computing, National University of Singapore and together with Jinlan Fu, are with Institute of Data Science, National University of Singapore, 117602. (E-mail: yangmt0418@gmail.com, seekiong@nus.edu.sg, and jinlanjonna@gmail.com)}
}
\markboth{Journal of \LaTeX\ Class Files,~Vol.~14, No.~8, August~2023}%
{Shell \MakeLowercase{\textit{et al.}}: A Sample Article Using IEEEtran.cls for IEEE Journals}

\maketitle

\begin{abstract}
Pre-trained conversation models (PCMs) have demonstrated remarkable results in task-oriented dialogue (TOD) systems. Many PCMs focus predominantly on dialogue management tasks like dialogue state tracking, dialogue generation tasks like response generation, or both. However, the existing PCMs seldom consider dialogue comprehension tasks, such as dialogue question answering and summarization tasks. These tasks allow PCMs to glean dialogue context from various angles. This observation naturally raises the question: Can the performance of downstream dialogue tasks be enhanced if a PCM is pre-trained on dialogue management, generation, and comprehension tasks?

To investigate this, we proposed an \textbf{Omni}potent \textbf{Dialog}ue pre-training model (OmniDialog). It unifies these three dialogue tasks into a monolithic framework by multi-task learning, fostering inter-task communication. The pre-training corpus of OmniDialog spans $\mathbf{7}$ dialogue-focused tasks, drawing from $\mathbf{15}$ datasets and encompassing over $\mathbf{3.2}$ million dialogue utterances.
To our knowledge, OmniDialog is a pioneering PCM pre-trained across dialogue management, generation, and comprehension domains. We evaluated its performance across four tasks: dialogue summarization, end-to-end dialogue modeling, dialogue state tracking, and intent classification. The results underscore its efficacy in domain transfer learning, low-resource, and full-dataset scenarios.
Furthermore, to glean a nuanced understanding of OmniDialog's strengths and potential pitfalls, we designed a fine-grained analysis framework for dialogue-centric tasks. Experimental results show that the OmniDialog is good at hard samples, such as long dialogues and lengthy responses. 
\end{abstract}


\section{Introduction}
Task-oriented dialogue (TOD) systems interactively help users achieve specific goals, such as booking tickets, through natural language. It can be decomposed into three subtasks~\cite{he2022unified}: (1) Dialog Understanding, (2) Policy Planning, and (3) Dialog Generation. 
Dialogue management~\cite{brabra2021dialogue}, also known as the dialogue policy, is an essential module of the TOD system, helping the system make decisions of optimal action. It selects system actions at each stage, enabling the dialogue to complete the goal successfully.

The rapid evolution of the pre-training language models, such as GPT-2~\cite{GPT-2}, T5~\cite{T5}, and GPT-3~\cite{brown2020language}, makes many tasks achieve new state-of-the-art performance. 
Many prior studies tried to build pre-trained conversation models (PCMs)~\cite{DialoGPT,bao2019plato,liu2021pretraining,he2022galaxy,henderson2019convert,su2021multi,he2022unified} by constructing pre-trained language models on extensive dialogue corpus. Then, the fine-tuning techniques are employed to adapt the PCM to downstream dialogue-related tasks, such as dialogue state tracking.
PCMs can be divided into three categories according to pre-training tasks they considered in the pre-training phase,  PCMs focus on  (1) dialogue generation tasks~\cite{DialoGPT,liu2021pretraining,he2022unified}, (2) dialogue management tasks~\cite{he2022galaxy,henderson2019convert}, and (3) both dialogue generation and dialogue management tasks~\cite{su2021multi,bao2019plato}.   
(1) For PCMs focusing on the dialogue generation tasks: DialoGPT~\cite{DialoGPT} with the GPT-2~\cite{GPT-2} backbone and TOP+NOD~\cite{liu2021pretraining} were pre-trained on the large-scale dialogue response generation (NLG) datasets collected from Reddit.
SPACE~\cite{he2022unified} is a unified semi-supervised pre-training model trained on a large corpus of dialogue response generation tasks. 
(2) For PCMs that focus on the dialogue management tasks:
GALAXY~\cite{he2022galaxy} focuses on policy optimization (i.e., end-to-end dialog modeling) by pre-training on a large dialogue corpus of dialog act prediction tasks;
ConveRT~\cite{henderson2019convert} was pre-trained on the response selection task built based on the Reddit conversational corpus. 
(3) For PCMs that focus on  both dialogue generation and dialogue management tasks:
PPTOD\cite{su2021multi} continually pre-trained a T5~\cite{T5} model on large human-system conversation datasets, covering dialogue state tracking (DST), dialogue policy learning (POL), intent classification (IC), and dialogue response generation (NLG).
PLATO~\cite{bao2019plato} focused on improving the performance of dialogue generation and pre-trained on both large datasets built based on the dialogue response generation (NLG) and action prediction tasks.

These previous methods based on building pre-trained conversation models significantly improve the performance of downstream dialogue tasks.
However, previous methods seldom consider the dialogue comprehension tasks, such as dialogue reading comprehension~\cite{cui2020mutual,sun2019dream} and dialogue summarization~\cite{chen2021dialogsum,zhao2021todsum} tasks in the pre-training phrase. Dialogue comprehension tasks make the PCMs learn the contextualized features of dialogue from different perspectives.

In this paper, we proposed an \textbf{Omni}potent \textbf{Dialog}ue pre-training model (OmniDialog), a multi-task pre-training framework~\cite{sanh2021multitask}, by pre-training on all-encompassing dialogue tasks, including the dialogue management, generation, and comprehension tasks. 
All considered pre-training tasks, covering $7$ tasks, $\mathbf{15}$ datasets, and over 3.2 million utterances in the dialogue domain.
What's more, to unify all the different format tasks into a monolithic pre-training framework, we carefully designed prompt templates~\cite{liu2023pre} for each task and converted them into the sequence-to-sequence format.
The pre-training tasks include response generation (NLP), dialogue state tracking (DST), dialogue policy learning (POL), intent classification (IC), next utterance prediction (NUP), dialogue multi-choice  question answering (MCQA), and dialogue summarization (SUMM).

We evaluate OmniDialog on 4 downstream tasks, namely end-to-end dialogue modeling~\cite{he2022unified}, dialogue state tracking~\cite{zhou2021dialogue}, intent classification~\cite{peng2020soloist}, and dialogue summarization~\cite{zhao2021todsum}.  Compared to previous state-of-the-art models \cite{su2021multi,liu2021pretraining,zhou2021dialogue} with comparable model size and experimental settings, OmniDialog achieves comprehensive and competitive performance in both full-dataset and low-resource settings. To further analyze the capability of the OmniDialog, we design a fine-grained analysis framework for dialogue-related tasks (e.g., DST). Experimental Results on four tasks demonstrated that our OmniDialog outperforms the baseline models on hard samples, such as long dialogue and response.

The main contributions of this work are listed as follows:

\begin{itemize}
    \item To the best of our knowledge, our work is pioneering in integrating dialogue comprehension tasks into a dialogue pre-training model.
    
    \item We proposed a fine-grained analysis framework for dialogue-related tasks, allowing for an examination of the model's strengths and weaknesses.

    \item Our experimental evaluation encompasses a wide range of tasks and scenarios, including domain transfer learning, full dataset, and low-resource settings. This comprehensive evaluation provides a thorough assessment of the model's performance in each scenario.

    \item To make different tasks interact better in a monolithic framework, we designed a set of prompts for each task that are participants in the OmniDialog pre-training.

\end{itemize}

\section{Related Work}
\textbf{Task-Oriented Dialogue.}
In Task-Oriented Dialogue (TOD), the dialogue system is designed to engage in goal-driven conversations with users, aiming to accomplish specific tasks or operations. The conversations typically revolve around tasks specific to a particular domain, such as product ordering, flight booking, or information retrieval. The system interacts with the user through dialogue to understand their needs, provide necessary information, and perform required operations to ultimately fulfill the user's goals. Traditional approaches process TOD as a pipeline  \cite{young2013pomdp}, involving dialogue state tracking, dialogue policy learning, and response generation. Recent methods increasingly leverage neural networks  \cite{liang2020moss, wen2016network} or Pre-trained Language Models (PLMs)  \cite{DialoGPT, su2021multi, he2022unified} to handle TOD tasks.

\textbf{Prompt Learning. } Pre-trained Language Models (PLMs) aim to leverage extensive text data to extract latent semantic information, enabling efficient processing of various NLP tasks. For downstream applications, PLMs are further fine-tuned using task-specific data. PLMs have emerged as a prominent approach in NLP, encompassing natural language understanding  \cite{peters2018deep,devlin2018bert} and text generation  \cite{BART, T5}. Recently, ChatGPT, whose backbone model is GPT-3.5, incorporates human feedback during training, exhibiting exceptional performance across diverse NLU and NLG tasks. In the domain of natural language processing, prompts have emerged as essential tools in guiding the output generation of language models~\cite{liu2023pre, wei2021finetuned,lester2021power, vu2021spot}. Prompts serve as initial instructions or cues provided to the model, shaping its response and controlling the generated text. They play a pivotal role in specifying the desired context, and desired output, or posing a specific question to direct the model's behavior. \cite{liu2023pre} Crafted with care, prompts enable users to elicit targeted information or style from the language model. They can vary in length and complexity, ranging from concise phrases to comprehensive paragraphs, depending on the desired level of guidance. The effective design of prompts empowers users to exert control over language models, allowing for the customization and shaping of the generated narrative or information in a desired manner.

\textbf{Pre-trained Dialogue Model.}
To help the model generate a dialogue-style response, many studies pre-trained the PLMs with dialogue-based corpora. DialoGPT \cite{DialoGPT}, which is pre-trained on the Reddit dataset, is designed solely for response generation and does not consider dialogue comprehension or dialogue management tasks. ConvBERT \cite{jiang2020convbert} used a novel span-based dynamic convolution to replace these self-attention heads in BERT, showing good performance in GLUE \cite{wang2018glue} and SQuAD \cite{rajpurkar2016squad} benchmarks. SOLOIST \cite{peng2020soloist} is another pre-trained model that focuses on DST and response generation data but lacks comprehension pre-training data. Nevertheless, it achieves excellent performance in few-shot settings. GALAXY \cite{he2022galaxy} combines task-oriented corpora into a new pre-training dataset called UniDA, primarily used for predicting dialogue acts, but it lacks comprehension pre-training tasks. PEGASUS  \cite{zhang2020pegasus}, a commonly used abstractive pre-training model, is pre-trained on C4 and HugeNews datasets without dedicated dialogue pre-training data. PPTOD \cite{su2021multi} introduced a unified plug-and-play dialogue model based on T5, showing good performance in end-to-end dialogue modeling, dialogue state tracking, and intent classification tasks.  SPACE \cite{he2022unified} introduces contrastive learning into pre-training but its NLU pre-training data does not include the summarization task.

\textbf{What is the difference between OmniDialog and existing Pre-training conversation models?}
The novelty of our OmniDialog mainly includes:
\begin{itemize}
    \item Previous researchers typically only consider two types of tasks: dialogue generation and dialogue management. In contrast, our OmniDialog, during the pre-training phase, not only considers dialogue generation and management but also incorporates multiple tasks related to dialogue comprehension. These tasks include dialogue summarization, multiple-choice QA, and NUP tasks.
    \item We have conducted a detailed study on the impact of introducing three dialogue comprehension tasks on the performance of several TOD-related tasks (such as DST).
    \item We have designed a fine-grained evaluation framework suitable for various tasks within the dialogue domain.
\end{itemize}

\section{Method}
\begin{figure*}[t]
  \centering
  \includegraphics[width=\linewidth]{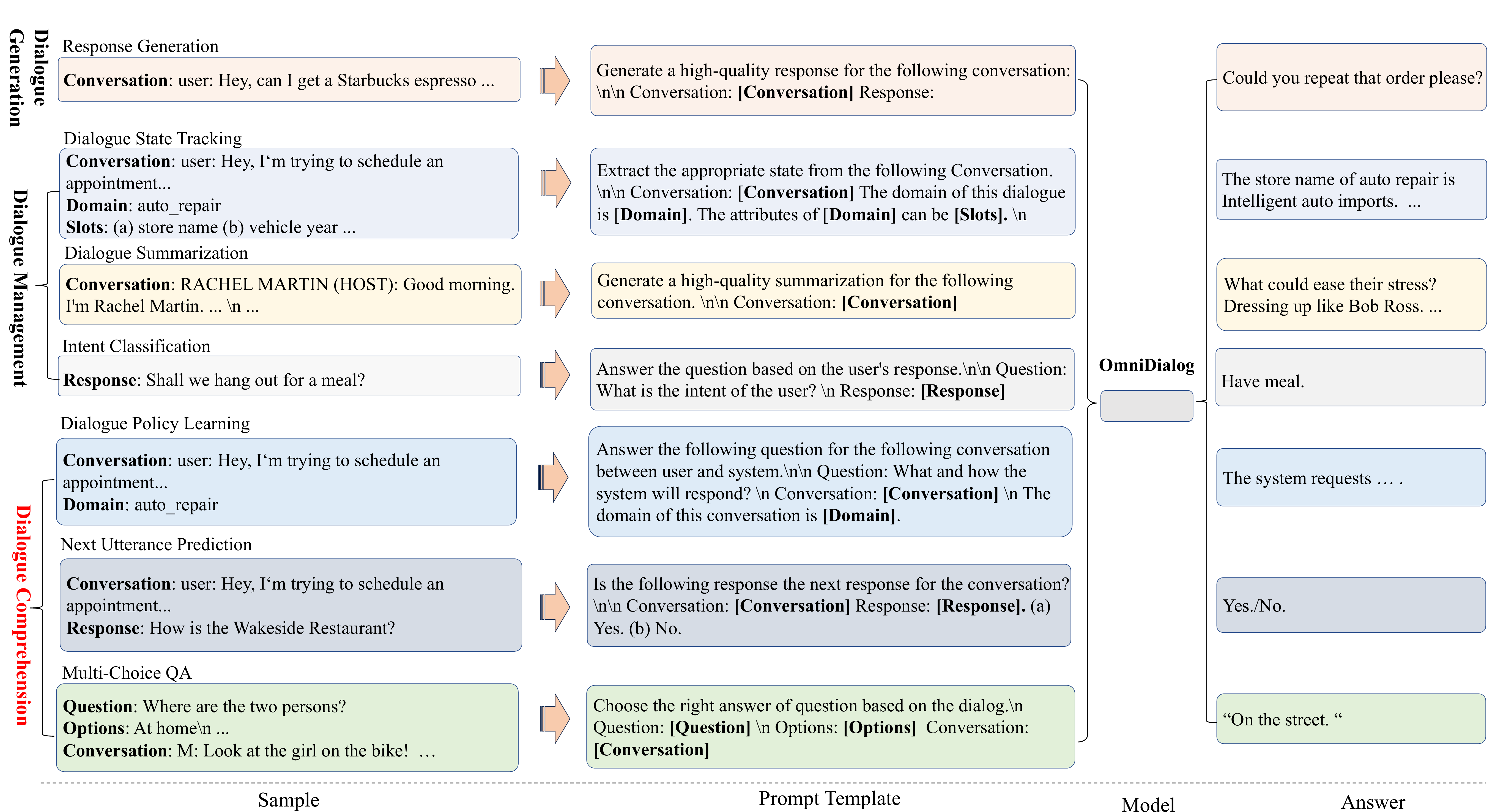}
  \caption{The framework of OmniDialog. It unifies seven distinct tasks from dialogue management, dialogue generation, and dialogue comprehension into a monistic pre-training model.}

\label{f1}
\end{figure*}

To train an omnipotent dialogue model that is proficient in multiple tasks in dialogue, we unify the $7$ tasks in the dialogue domain into a monolithic framework, named OmniDialog. These $7$ tasks cover dialogue management, dialogue generation, and dialogue comprehension tasks. 

Fig. \ref{f1} shows the framework of OmniDialog.
In this section, first, we will introduce the pre-training datasets; second, the OmniDialog pre-training framework; third, the prompt design for the $7$ pre-training tasks.

\subsection{Pre-training Tasks and Datasets}

\begin{table}[h]
  \caption{Statistics of the Pre-training Datasets. Ge. represents the Dialogue Generation task. Co. represents the Dialogue Comprehension task. Ma. represents the Dialogue Management task.}

 \resizebox{\columnwidth}{!}{%
  \begin{tabular}{llccc}
    \toprule
    Dataset & Tasks &Task Type& No. of Utterances & No. of Domains \\
    \midrule
    MetaLWOZ \cite{lee2019multi} & NLG &Ge.& 822,932 & 47\\
    MuTual \cite{cui2020mutual} & QA &Co.& 41,908 & Open-domain\\
    DREAM \cite{sun2019dream} & QA&Co. & 30,183 &  Open-domain\\
    SNIPS \cite{coucke2018snips} & IC&Co. & 25,682 & 9\\
    CLINC \cite{larson2019evaluation} & IC&Co. & 45,000 & 10\\
    ATIS \cite{ATIS} & IC&Co. & 10,772 & 1\\
    Ubuntu \cite{lowe2015ubuntu} & NUP&Co. & 7,100,000 & 1\\
    MediaSum \cite{zhu2021mediasum} & SUMM & Co. & 463,596 & Open-domain\\
    KVRET \cite{eric2017key} & DST, NLG&Ge., Ma. & 31,504 & 3\\
    WOZ \cite{mrkvsic2016neural} & DST, NLG&Ge., Ma. & 15,248 & 1\\
    TaskMaster \cite{byrne2019taskmaster} & DST, NLG&Ge., Ma. & 540,688 & 6\\
    CamRest676 \cite{wen2016network} & DST, NLG&Ge., Ma. & 10,976 & 1\\
    MSR-E2E \cite{li2018microsoft} & DST, POL, NLG&Ge., Ma. & 72,238 & 3\\
    Frames \cite{asri2017frames} & DST, POL, NLG&Ge., Ma. & 38,316 & 1\\
    Schema-Guided \cite{meyer2003library} & DST, POL, NLG&Ge., Ma. &757,380 & 17\\
  \bottomrule
\end{tabular}
}
\label{table2}
\end{table}

We constructed the pre-training corpus by collecting existing datasets in the dialogue domain. Tab. \ref{table2} shows the statistical information of the pre-training datasets.
It covers 7 tasks, including Dialog State Tracking (DST),  Dialogue Policy Learning (POL), Natural Language Generation (NLG), Intent Classification (IC), Dialogue Multiple Choice Question Answering (MCQA), Next Utterance Prediction (NUP), and Dialogue Summarization (SUMM).
Notably, we select 50,000 and 90,000 samples from the Ubuntu and Media datasets, respectively. 
Ultimately, we constructed a pre-training corpus that encompasses 7 tasks, covers 15 datasets, and includes 3.2 million training samples.

\subsection{OmniDialog: Omnipotent Dialogue Pre-training Model}

Inspired by previous works~\cite{su2021multi, T5,worsham2020multi} that consolidate various NLP tasks into a single pre-trained model, we aim to incorporate as many tasks from the dialogue domain as possible. This integration is designed to enhance the model's understanding of the unique features and intrinsic interactions of tasks within the dialogue domain. For each sample, we augment the raw input with a prompt specific to the task.
Given the prompt $z_t$ for task $t$, each training sample $d_t$ in task $t$ can be represented as:
\begin{equation}
  \hat{x}, \hat{y} = z_t(x, y),
\end{equation}
where $x$ and $y$ denote the plain text of the input and output before equipping with the prompt template. 
Here, the task \textit{t} can be  \textit{response generation (NLG)}, \textit{dialogue state tracking (DST)}, \textit{dialogue policy learning (POL)}, \textit{intent classification (IC)}, \textit{multi-choice question answering (MCQA)}, \textit{next utterance prediction (NUP)} or \textit{dialogue summarization (SUMM)}. 

\textbf{Learning.} As for learning, we select the maximum likelihood objective. Given the training sample ($\hat{x}$, $\hat{y}$) that equipped with prompt $z_t$ of task $t$, the learning objective $L_{\theta}$ is defined as
\begin{equation}
  L_{\theta}=-\sum_{i=1}^{|\hat{y}|} \log(P_{\theta}(\hat{y}_{i}|\hat{y}_{<i};\hat{x},z_t)),
\label{eq2}
\end{equation}
in which $\theta$ represents the model parameters.

\subsection{Prompt Design}
Prompt engineering techniques \cite{liu2023pre} help machines better understand and interact with humans. Generally speaking, it can be divided into two categories: hard prompt \cite{liu2023pre, wei2021finetuned} and soft prompt \cite{lester2021power, vu2021spot}. The hard prompt is hand-crafted and interpretable natural language texts, while the soft prompt is a tunable feature vector that can be optimized by training.
In this work, we carefully design a set of prompts for the seven tasks that OmniDialog considered, which provides enough task information and hints. We show the prompt template for the pre-training task in Fig. \ref{f1}.

\section{Experiments and Results}
\label{sec:exp}
To evaluate the performance of our model, we tested the OmniDialog on $5$ datasets covering four tasks: (1) namely End-to-End Dialogue Modelling (E2E); (2) Dialog State Tracking (DST); (3) Intent Classification (IC);  and (4) Dialogue Summarization (SUMM).

\subsection{Experiemntal Settings}

\subsubsection{Evaluation Tasks and Datasets}
The evaluation tasks and datasets considered in this work are listed below:

\noindent
\paragraph{End-to-End Dialogue Modelling}~ \cite{hosseini2020simple} refers to predicting the system response from a user's speech or text input in a conversational system. Here, we consider the MultiWOZ 2.0 \cite{budzianowski2018multiwoz} benchmark, which is a dataset for TOD systems. In typical MultiWOZ 2.0 works \cite{budzianowski2018multiwoz}, the generated responses are supported by both dialog history and a database (DB) state. We follow the generation process of PPTOD \cite{su2021multi}.

\noindent
\paragraph{Dialog State Tracking} aims to track and update the state of a dialog system as a user interacts with it. 
Here, we consider the MultiWOZ 2.0 benchmark~ \cite{budzianowski2018multiwoz}.
 
\noindent
\paragraph{Intent Classification} aims to detect the user's intent, given the user's response. Here, the Banking77 benchmark~ \cite{casanueva2020efficient} with 77 banking scenarios is considered. 

\noindent
\paragraph{Dialogue Summarization} aims to extract key information and generate concise summaries from dialogues. In this work, we select TODSum \cite{zhao2021todsum} and DialogSum \cite{chen2021dialogsum} as our evaluation benchmarks.

\subsubsection{Evaluation Metric}
\paragraph{Combined Score}~\cite{budzianowski2018multiwoz} is broadly selected as the evaluation metric for the E2E dialogue modeling task.
It can be formalized as: 
\begin{equation}
  \text{Combined \ Score} = \text{BLEU} + 0.5 \times (\text{Inform Rate} + \text{Success Rate}),
\label{eq3}
\end{equation}
where BLEU \cite{papineni2002bleu} is a typical metric reflecting the similarity between the generated responses with the target responses. Inform Rate \cite{budzianowski2018multiwoz} is to reflect whether the generated response offers an appropriate entity, while Success Rate \cite{budzianowski2018multiwoz} reflects whether the response answers all the requested attributes(e.g., restaurant address).

\paragraph{Joint Goal Accuracy (JGA)}~\cite{henderson2014second} is used as the evaluation metric of the DST task. It measures the percentage of successful task completions in which all participants in the dialogue reach the same goal.

\paragraph{ROUGE Score}
We choose ROUGE  \cite{lin2004rouge} as the evaluation metric for the dialogue summarization task. Specifically, 
the three variants of the ROUGE are utilized, namely ROUGE-1 (R1), ROUGE-2 (R2), and ROUGE-Lare (RL).

\subsubsection{Baselines}
\label{sec:baselines}

We compared our model with existing popular methods. 

\noindent
\paragraph{E2E Task: } the main baselines include DAMD \cite{zhang2020task}, MinTL \cite{lin2020mintl},  DoTS \cite{jeon2021domain}, DORA \cite{jeon2022dora}, SimpleTOD \cite{hosseini2020simple}, UBAR \cite{yang2021ubar}, JOUST \cite{tseng2021transferable}, SOLOIST \cite{peng2020soloist}, TOP and TOP+Noisy Online Decoding (TOP+NOD) \cite{liu2021pretraining},  LAVA \cite{lubis2020lava}, and PPTOD \cite{su2021multi}.

\noindent
\paragraph{DST Task: } the considered baselines are TRADE \cite{wu2019transferable}, COMER \cite{ren2019scalable}, DSTQA \cite{zhou2019multi}, SOM-DST \cite{kim2019efficient}, MinTL \cite{lin2020mintl}, SOLOIST \cite{peng2020soloist}, UBAR \cite{yang2021ubar},  and PPTOD \cite{su2021multi}.

\noindent
\paragraph{Intent Classification Task: } the selected main baselines are BERT-Fixed \cite{casanueva2020efficient}, BERT-Tuned \cite{casanueva2020efficient}, USE+ConveRT \cite{casanueva2020efficient}, USE \cite{yang2019multilingual},  SOLOIST \cite{peng2020soloist}, TOD-BERT \cite{wu2020tod}, ConveRT \cite{henderson2019convert}, and PPTOD \cite{su2021multi}.

\noindent
\paragraph{Dialogue Summarization: } The considered baselines include 
Lead-3 \cite{see2017get}, Lead-3 \cite{see2017get},
BertExt \cite{liu2019text},
Oracle \cite{narayan2018don},
BertAbs w. DS(oracle) \cite{zhao2021todsum},
BART w. DS(oracle) \cite{zhao2021todsum},
LONGEST \cite{gliwa2019samsum}, 
EXT-ORACLE \cite{narayan2018don},
DistilBART \cite{shleifer2020pre},
PEGASUS \cite{zhang2020pegasus}.

\noindent
\paragraph{Variants of OmniDialog: }
To examine the effectiveness of introducing dialogue comprehension tasks, that is, MCQA (M), NUP (N), and Summ (S), we considered four variant models of OmniDialog.
\begin{itemize}
    \item \textbf{OmniDialog}: its pre-training datasets include all datasets for dialogue management, generation, and comprehension tasks.
    \item \textbf{OmniDialog-MNS}: This model's pre-training datasets include all the datasets for dialogue management and generation; however, the datasets for the dialogue comprehension tasks (e.i., MCQA (M), NUP (N), and Summ (S)) are excluded.
    \item  \textbf{OmniDialog-MN}: The pre-training datasets of this model encompass all datasets except the MCQA and NUP tasks, which belong to the dialogue comprehension tasks.
    \item \textbf{OmniDialog-S}: The pre-training datasets of this model encompass all datasets except the text summarization dataset, which belongs to the dialogue comprehension tasks.
\end{itemize}

\subsubsection{Implementation Details}
Our model OmniDialog and its variants are initialized on T5-base (OmniDialog \textsubscript{base}), whose numbers of parameters are about 220M. 
We also study our OmniDialog on a small model which is based on the T5-small.

\paragraph{Pre-training phrase:} We train our models for 5 epochs. The maximal sequence length is 1000.  We select Adam as our optimizer \cite{kingma2014adam} with a learning rate of 5e-5 and a batch size of 128.

\paragraph{Fine-tuning phrase:} The batch size for E2E dialogue modeling, DST, and IC is 128, and the learning rate is 5e-4. The batch size for dialogue summarization is 32 and the learning rate is 3e-5. All the work is implemented with the Huggingface Library \cite{wolf2019huggingface}.

\subsection{End-to-End Dialogue Modeling}

\begin{table}[htp]
  \caption{End-to-end Modeling Results. The values in bold in each column are the best performers for this metric. OmniDialog-MNS, OmniDialog-MN, and OmniDialog-S are variants of OmniDialog based on T5-base. 
  }
  \label{tab:freq}
  \resizebox{\columnwidth}{!}{%
  \begin{tabular}{lcccc}
    \toprule
    Model & Inform & Success & BLEU & Combined Score \\
    \midrule
    DAMD \cite{zhang2020task} & 76.33 & 60.40 & 16.60 & 84.97 \\
    SimpleTOD \cite{hosseini2020simple} & 84.40 & 70.10 & 15.01 &92.26 \\
    UBAR \cite{yang2021ubar} & 85.10 & 71.02 &16.21 & 94.27\\
    DoTS \cite{jeon2021domain} & 86.59 & 74.14 & 15.06 & 95.43\\
    DORA \cite{jeon2022dora} &  85.60 &  74.60 &  15.35 & 95.45\\
    % HIER-Joint \cite{santra2020hierarchical} & 80.50 & 71.70 & 19.74 & 95.84\\
    SOLOIST \cite{hosseini2020simple} & 85.50 & 72.90 & 16.54 & 95.74\\
    JOUST \cite{tseng2021transferable} & 83.20 & 73.50 & 17.60 & 95.95\\
    TOP \cite{liu2021pretraining} & 85.20 & 72.90 & 17.00 & 96.05 \\
    MinTL \cite{lin2020mintl} & 84.88 & 74.91 & 17.89 & 97.78 \\
    LAVA \cite{lubis2020lava} & 91.80& \textbf{81.80}& 12.03 & 98.83 \\
    PPTOD \textsubscript{small} \cite{su2021multi} & 87.80 & 75.30 & 19.89 & 101.44\\
    TOP+NOD \cite{liu2021pretraining} & 86.90 & 76.20 & \textbf{20.58} & 102.13\\
    PPTOD \textsubscript{base} \cite{su2021multi} & 89.20 & 79.40 & 18.62 & 102.92\\
    \midrule
    OmniDialog & 92.20 & 79.30 & 18.57 & \textbf{104.32}\\
    OmniDialog-MNS & \textbf{92.60} & 75.70 & 16.35 & 100.50 \\
    OmniDialog-MN & 90.10 & 76.90 & 18.14 & 101.64 \\
    OmniDialog-S & 89.90 & 73.40 & 17.31 & 98.96 \\

  \bottomrule
\end{tabular}}
\label{table3}
\end{table}

Tab. \ref{table3} shows the results of the end-to-end dialogue modeling task on the MultiWOZ 2.0 dataset. 
The main observations are summarized below:

1. \textit{OmniDialog outperforms most previous state-of-the-art methods on the \texttt{combine score}.} The reason can be attributed to the carefully designed prompt templates for different tasks. Our model achieves a relatively high \texttt{inform score}, meaning that the responses generated by our model show distinguished ability in offering appropriate entities. Additionally, we have observed a decrease in the BLEU score, which could potentially be attributed to the fact that the pre-training data includes a wide range of outputs, consequently deviating generated responses from the intended target responses.  

2. \textit{Incorporating dialogue comprehension tasks significantly improved the model's performance.} We observed that the performance of OmniDialog, which is pre-training on dialogue comprehension tasks (e.i., MCQA, NUP, and Summ), outperforms the OmniDialog-MNS without considering the dialogue comprehension tasks by 3.82\%.

3. \textit{The introduction of text summarization task is beneficial for End-to-End (E2E) dialogue modeling significantly.} 
Comparing the performance of the variants, we discovered that the models pre-trained on text summarization tasks, namely OmniDialog and OmniDialog-MN, both outperform the models that do not incorporate summarization tasks, which are OmniDialog-MNS and OmniDialog-S. This suggest that the inclusion of supplementary summarization data contributes to a deeper comprehension of dialogue history.

\subsection{Dialogue State Tracking}

\begin{table}[htp]
  \caption{Results of Dialogue State Tracking. The values in bold in each column are the best performers for this metric.}
  \centering
  \label{tab:ab-d}
  % \resizebox{\columnwidth}{!}{%
  \begin{tabular}{lc}
    \toprule
    Model & JGA \\
    \midrule
    TRADE \cite{wu2019transferable}  & 48.62 \\
    COMER \cite{ren2019scalable} & 48.79 \\
    DSTQA \cite{zhou2019multi} & 51.44 \\
    SOM-DST \cite{kim2019efficient} & 51.38 \\
    MinTL \cite{lin2020mintl} & 52.10 \\
    UBAR \cite{yang2021ubar} & 52.59 \\
    SOLOIST \cite{peng2020soloist} & 53.20 \\
    PPTOD \textsubscript{small} \cite{su2021multi} & 51.50 \\
    PPTOD \textsubscript{base} \cite{su2021multi} & 53.37 \\
    \midrule
    OmniDialog & \textbf{53.98}\\
    OmniDialog-MNS & 53.30 \\
    OmniDialog-MN & 53.81 \\
    OmniDialog-S & 52.60 \\

  \bottomrule
\end{tabular}
%}
\label{table4}
\end{table}

Tab. \ref{table4} showcases the outcomes of Dialogue State Tracking (DST) utilizing the MultiWOZ 2.0 dataset. 
The main findings are listed below: 

1. OmniDialog and OmniDialog-MN outperform all previous state-of-the-art models. This achievement underscores the value of incorporating comprehension pre-training data in facilitating the extraction of both intent and pertinent information. Moreover, even OmniDialog-MNS, which lacks comprehension data during pre-training, surpasses the majority of prior approaches. This outcome suggests that the devised prompts, enriched with domain and slot information for the corresponding domain, empower OmniDialog to more precisely extract the necessary value information from dialogues in a targeted manner.

2. As for the variants of OmniDialog, the outcomes from both end-to-end dialogue modeling and dialogue state tracking (DST) exhibit a consistent trend. 
OmniDialog-MN surpasses OmniDialog-MNS and OmniDialog-S, which indicate that the introduction of text summarization task enhances the understanding of dialogue history.

\subsection{Intent Classification}

\begin{table}[htpb]\centering
\caption{Results of Intent Classification. The values in bold in each column are the best performers for this metric.}
\begin{tabular}{lc}\toprule
Model &Accuracy \\\midrule
Bert-fixed \cite{casanueva2020efficient} &87.19 \\
Bert-tuned \cite{casanueva2020efficient} &93.66 \\
USE \cite{yang2019multilingual} &92.81 \\
ConveRT \cite{henderson2019convert} &93.01 \\
USE+ConveRT \cite{casanueva2020efficient} &93.36 \\
SOLOIST \cite{peng2020soloist} &93.80 \\
PPTOD \textsubscript{small} \cite{su2021multi} &93.24 \\
PPTOD \textsubscript{base} \cite{su2021multi} &93.76 \\
\midrule
OmniDialog & \textbf{93.86} \\
OmniDialog-MNS & 93.60 \\
OmniDialog-MN & 93.71 \\
OmniDialog-S & 93.69 \\
\bottomrule
\end{tabular}
\label{table5}
\end{table}

The performance of the intent classification on Banking77 dataset is shown in Tab. \ref{table5}. 
Here, we summarize the main observations as follows:

1. The accuracy of OmniDialog is higher than the previous SOTA (i.e., and PPTOD \textsubscript{base}) on the comparable model size. The tasks considered by the PPTOD only belong to dialogue management and generation, while OmniDialog is considered an extra type of task, namely dialogue comprehension, which makes our model stronger dialogue understanding ability.

2. As for the performance of the variants on intent classification, it is noteworthy that both OmniDialog-S and OmniDialog-MN demonstrate superior performance compared to OmniDialog-MNS. This suggests that the introduction of separate comprehension and summarization data aids the model in effectively discerning the conversational intent. Across the four models, OmniDialog emerges as the top performer, highlighting that the amalgamation of comprehension and summarization data further elevates the model's proficiency in intent identification.

\subsection{Dialogue Summarization}
\begin{table}[htpb]\centering
\caption{Results of dialogue summarization task on DialogSum dataset.}
\begin{tabular}{lccc}\toprule
Model & R1 & R2 & RL\\\midrule
LONGEST-2 \cite{gliwa2019samsum}&  24.15 & 6.25 & 22.73 \\
LEAD-2 \cite{see2017get} & 27.52 & 6.78 & 27.31 \\
EXT-ORACLE-2 \cite{extoracle}  & 37.90 & 13.88 & 34.04\\
Transformer \cite{vaswani2017attention} & 35.91 & 8.74 & 33.50 \\
DistilBART \cite{shleifer2020pre} & 35.93 & 11.71 & 28.86 \\
PEGASUS \cite{zhang2020pegasus} & 38.40 & 13.84 & 33.41 \\
\midrule
OmniDialog & \textbf{41.53} & \textbf{15.40} & \textbf{38.74} \\
OmniDialog-MNS &  38.85 & 13.98 & 36.52 \\
OmniDialog-MN  & 40.60 & 14.62 & 37.72\\
OmniDialog-S & 39.63 & 14.29 & 37.16 \\

\bottomrule
\end{tabular}
\label{table6}
\end{table}

The ROUGE scores for Dialogue Summarization task on the DialogSum and TODSum datasets are shown in Tab.~\ref{table6} and Tab.~\ref{table7}, respectively.
Below, we list the main conclusions:

1. OmniDialog outperforms the baselines, which holds for both DialogSum and TODSum datasets.
This result suggests that multi-task pre-training with other dialogue data enhances the performance of dialogue summarization, in contrast to PEGASUS (baseline on the DialogSum), which was exclusively pre-trained using summarization data. 

2. In the case of DialogSum, both OmniDialog-S and OmniDialog-MN exhibit notable superiority over OmniDialog-MNS, with OmniDialog leading the pack. This outcome underscores the positive impact of integrating both comprehension and summarization data, culminating in the production of high-quality summaries.

3. On the TODSum dataset, OmniDialog achieves the highest ROUGE scores compared to previous methods. Compared to the previous state-of-the-art method, BART with DS (oracle), despite not explicitly including the dialogue state (DS) in the input, OmniDialog surpasses BART with DS (oracle), indicating that OmniDialog can implicitly extract DS from the dialogue. This demonstrates OmniDialog's ability to outperform models that require the explicit inclusion of the DS in the input.

\begin{table}[htpb]\centering
\caption{Results of TODSum}
\begin{tabular}{lcccc}\toprule
Model  & R1 & R2 & RL\\\midrule
LEAD-3 \cite{see2017get} & 22.43 & 4.52 & 18.85 \\
BertExt \cite{liu2019text} & 42.10 & 12.45 & 36.55 \\
Oracle \cite{liu2019text}  & 44.90 & 14.24 & 38.69\\
BertAbs w. DS(oracle) \cite{zhao2021todsum} & 80.39 & 62.60 & 78.07\\
BART w. DS(oracle) \cite{zhao2021todsum} & 81.12 & 67.87 & 78.94\\
\midrule
OmniDialog & 83.85 & 73.19 & 82.99 \\
OmniDialog-MNS & 84.26 & 73.52 & 83.38 \\
OmniDialog-MN  & 84.22 & \textbf{74.43} & 83.28\\
OmniDialog-S & \textbf{84.67} & 74.21 & \textbf{83.84} \\
\bottomrule
\end{tabular}
\label{table7}
\end{table}

\section{Analysis}

\subsection{Low-Resource Evaluation}
To evaluate the generalization ability, we conduct low-resource evaluations on E2E Dialogue Modeling, Dialogue State Tracking (DST), and Intent Classification (IC). 

\noindent
\textbf{Datasets and Settings.} We select the same datasets as the full training explored in Sec.~\ref{sec:exp}. For E2E dialogue modeling and DST task, MultiWOZ 2.0 serves as the fine-tuning dataset, and Banking 77 dataset are studied for the IC task. 
As for the low-resource setting, varying the percentage of the training dataset, including 1\%, 5\%, 10\%, and 20\% are selected as the low-resource training datasets. Regarding IC, we vary the number of training samples labeled as each intent from 10 to 30 on the Banking77 dataset.

\begin{table*}[htp]
\centering
\caption{Low-resource Evaluation for E2E Dialogue Modeling on MultiWOZ 2.0 dataset.}
\label{table9}
\resizebox{\textwidth}{!}{%
\begin{tabular}{l|cccc|cccc|cccc|cccc}
\toprule
\multirow{2}{*}{Model} & 
\multicolumn{4}{c|}{1\% of training data} &
\multicolumn{4}{c|}{5\% of training data} &
\multicolumn{4}{c|}{10\% of training data} &
\multicolumn{4}{c}{20\% of training data} \\
\cline{2-17}
    & Inform & Succ. & BLEU & Comb. & Inform & Succ. & BLEU & Comb. & Inform & Succ. & BLEU & Comb. & Inform & Succ. & BLEU & Comb.\\
\hline
MD-Sequicity \cite{zhang2020task} & - & - & - & - & 49.40 & 19.70 & 10.30 & 44.85 & 58.10 & 34.70 & 11.40 & 57.80 & 64.40 & 42.10 & 13.00 & 66.25 \\
DAMD \cite{zhang2020task} & 34.40 & 9.10 & 8.10 & 29.85 & 52.50 & 31.80 & 11.60 &53.75& 55.30 &30.30& 13.00 &55.80 &62.60& 44.10& 14.90 &68.25\\
SOLOIST \cite{hosseini2020simple}& 58.40 &35.30 &10.58 &57.43& 69.30& 52.30& 11.80&72.60& 69.90 &51.90& 14.60& 75.50 &74.00& 60.10 &15.25& 82.29\\
MinTL \cite{lin2020mintl} &-& -& -& - &75.48 &60.96 &13.98& 82.20& 78.08 &66.87& 15.46& 87.94 &82.48 &68.57& 13.00& 88.53\\
\hline
PPTOD \textsubscript{small} \cite{su2021multi}& 66.96& 50.90& 12.51 &71.44& 76.58 &61.60 &\textbf{15.35} &84.44 &83.50 &68.18 &15.56 &91.01& 82.96 &69.90 &17.02 &93.45\\
PPTOD \textsubscript{base} \cite{su2021multi}& 74.42& 52.44& \textbf{12.99}& 76.41& 79.86& 63.48 &14.89 &86.55 &84.42& 68.36& 15.57 &91.96& 84.94& 71.70 &17.01 &95.32\\
\hline
OmniDialog \textsubscript{small}  & 76.30 & 50.10 & 10.78&73.98 &\textbf{93.50} & 67.50 & 14.28 & 89.78 & 84.10 & 68.30 & 15.52 & 91.72 & \textbf{87.99} & 68.80 & 16.73 & 95.13 \\
OmniDialog \textsubscript{base} & \textbf{86.20} & \textbf{64.60} & 10.94 & \textbf{86.34} & 86.60 & \textbf{77.40} & 14.14 & \textbf{96.14} & \textbf{86.80} & \textbf{70.60} & \textbf{15.59} & \textbf{94.29} & 83.80 & \textbf{73.70} & \textbf{18.12} & \textbf{96.87}\\
\bottomrule
\end{tabular}
}
\end{table*}

\noindent
\textbf{Results and Analysis. } The results for E2E Dialogue Modeling, Dialogue State Tracking, and Intent Classification are shown in Tab.~\ref{table9}, Tab.~\ref{table10}, and Tab.~\ref{table11}, respectively.
Below we list the main observations and conclusions from these three tasks respectively.

\textbf{1. E2E Dialogue Modeling (Tab.~\ref{table9}).} OmniDialog consistently outperforms the other baseline models across different resource settings. Moreover, the combined scores of OmniDialog trained with 20\% of the training data are sometimes higher than those of previous SOTAs trained with the full dataset. 

\begin{table}[htp]
\centering
\caption{Low-resource Evaluation for DST on MultiWOZ 2.0 dataset.}
\label{table10}
\begin{tabular}{lcccc}
\toprule
\multirow{2}{*}{Model} & \multicolumn{4}{c}{Training Data Percentage} \\
\cline{2-5}
& 1\% & 5\% & 10\% & 20\% \\
\hline
SimpleTOD  \cite{hosseini2020simple} & 7.91& 16.14 &22.37& 31.22 \\
MinTL  \cite{lin2020mintl}& 9.25& 21.28& 30.32& 35.96\\
SOLOIST  \cite{peng2020soloist}&  13.21&  26.53&  32.42&  38.68\\
\hline
PPTOD \textsubscript{small}  \cite{su2021multi} &27.85& 39.07& 42.36& 45.98\\
PPTOD \textsubscript{base}  \cite{su2021multi} & 29.72& 40.20& 43.45& 46.96\\
\hline
OmniDialog \textsubscript{small} & 26.92 & 41.93 & 44.46 & 49.00\\
OmniDialog \textsubscript{base} & \textbf{32.73} & \textbf{42.12} &\textbf{45.49} & \textbf{49.50}\\

\bottomrule
\end{tabular}
\end{table}

\textbf{2. Dialogue State Tracking (Tab. \ref{table10}).} The OmniDialog \textsubscript{base} achieves the best performance. 
The reasons can be attributed to the well-designed prompts, which make the communication between different tasks much more easier.
In the case of low-resource with 20\% of the training dataset, OmniDialog \textsubscript{base} even beats some methods that were trained with the whole dataset.

\begin{table}[htp]
\centering
\caption{Low-resource Evaluation for IC task on Banking77 dataset.}
\label{table11}
\begin{tabular}{lccc}
\toprule
\multirow{2}{*}{Model} & \multicolumn{3}{c}{Number of Training Samples} \\
\cline{2-4}
& 10 & 30 & Full \\
\hline
BERT-Fixed  \cite{casanueva2020efficient} & 67.55 & 80.07 & 87.19 \\
BERT-Tuned  \cite{casanueva2020efficient} & 83.42 & 90.03 & 93.66 \\
USE  \cite{yang2019multilingual} & \textbf{84.23} & 89.74 & 92.81 \\
ConveRT  \cite{henderson2019convert} & 83.32 & 89.37 & 93.01 \\
% USE+ConveRT  \cite{casanueva2020efficient} & 85.19 & 90.57 & 93.36 \\
SOLOIST  \cite{peng2020soloist} & 78.73 & 89.28 & 93.80 \\
PPTOD \textsubscript{small}  \cite{su2021multi} & 78.87 & 87.88 & 93.27 \\
PPTOD \textsubscript{base}  \cite{su2021multi} & 82.81 & 89.64 & \textbf{93.86} \\
\hline
OmniDialog \textsubscript{small} & 77.46 & 88.24 & 93.14 \\
OmniDialog \textsubscript{base} & 83.34 & \textbf{90.16} & \textbf{93.86} \\
\bottomrule
\end{tabular}
\end{table}

\textbf{3. Intent Classification (Tab. \ref{table11}). } OmniDialog outperforms all previous SOTAs when fine-tuned with 30 samples per intent. This suggests that incorporating supplementary datasets for dialogue comprehension enhances the model's generality, thereby necessitating an ample amount of data to achieve improved performance in Intent Classification.

\subsection{Domain Transfer Learning}
To further explore the generalization ability of OmniDialog, we evaluated domain adaptation in the dialogue summarization task on the TODSum dataset~\cite{zhao2021todsum}. 

\paragraph{\textbf{Experimental Settings}} 
The TODSum dataset comprises five domains: train, taxi, restaurant, hotel, and attraction. 
Like previous approaches~\cite{zhao2022domain,zhao2022adpl}, we focus on the dialogues belonging to a single domain and exclude the dialogues included in multiple domains.

Dialogues from four out of the five domains are combined to create the source domains, while the remaining domain serves as the target domain. 200 source domain samples are randomly selected from source domains to build the validation set.

The baseline models include: Lead-3   \cite{see2017get}, 
Oracle   \cite{liu2019text}, BertExt   \cite{liu2019text}, BertAbs   \cite{liu2019text}, 
PGN   \cite{see2017get}, Transformer   \cite{vaswani2017attention},  BART   \cite{BART}, 
M-BART   \cite{chen2021structure}, 
BART w. DS   \cite{zhao2021todsum}, 
Prefix-tuning (BART)   \cite{zhao2022adpl}, Pegasus   \cite{zhang2020pegasus}, 
Prefix-tuning (Pegasus)   \cite{zhao2022adpl}, DOP   \cite{zhao2022domain}, and ADPL   \cite{zhao2022adpl}.

\begin{table}[htp]\centering
\caption{ROUGE-L Scores of Domain Adaptation on TODSum. The values in bold in each column are the best results. \textit{Resta.} and \textit{Attra.} denote \textit{Restaurant} and \textit{Attraction} domain, respectively. $\ast$: The backbone of DOP is BART-large. $\dagger$: The backbone of ADPL is Pegasus-large.}
\label{table8}
\resizebox{\columnwidth}{!}{%
\begin{tabular}{lccccc}
\toprule
Domain  & Taxi & Resta. & Attra. & Hotel& Train  \\
\hline
Lead-3   \cite{see2017get}  &20.75 &23.49 &19.66&18.80   &16.07\\
Oracle   \cite{liu2019text}    & 33.43  & 38.42& 38.79  & 32.56   & 32.87\\
BertExt   \cite{liu2019text}    & 33.36  & 34.43& 31.41  & 30.10   & 33.24\\
\hline
PGN   \cite{see2017get}    &29.82  &31.47  & 30.29 &30.93   &29.33\\
Transformer   \cite{vaswani2017attention}    &30.57  &31.99 &30.91 &31.63   &30.28\\
\hline
BertAbs   \cite{liu2019text}    &32.15  &38.87 &34.67 &33.22   &37.32\\
BART   \cite{BART}    &34.41  &44.93 &41.44 &36.83   &42.06\\
M-BART   \cite{chen2021structure}    &36.03  &45.00 &50.46 &38.23   &43.90\\
BART w. DS   \cite{zhao2021todsum}   &38.65  &45.23  &45.16 &39.31  &44.59\\
Prefix-tuning (BART)   \cite{zhao2022adpl}    &39.62  &42.99 &40.94 &36.75   &41.06\\
Prefix-tuning (Pegasus)   \cite{zhao2022adpl}    &40.71  &50.56 &47.63 &41.75   &45.56\\
Pegasus   \cite{zhang2020pegasus}   &43.34  &50.18 &50.96 & 42.17  &47.67\\
DOP   \cite{zhao2022domain}$\ast$    &42.75  &47.44 &49.48 &41.45   & 47.78\\
ADPL   \cite{zhao2022adpl}$\dagger$ & 45.62 & \textbf{56.37} & \textbf{51.49} & \textbf{45.16} & \textbf{52.36}\\
\hline
OmniDialog  \textsubscript{base}    & \textbf{46.09}  & 53.53 & 51.02 & 40.25   & 45.01 \\
\bottomrule
\end{tabular}
}
\end{table}

\paragraph{\textbf{Observation}} 
The results of zero-shot domain adaptation on TODSum are shown in Tab. \ref{table8}. The main observations are summarized below:
\begin{itemize}
    \item[(1)] OmniDialog demonstrates state-of-the-art (SOTA) performance in the restaurant, attraction, and taxi domains while exhibiting relatively lower performance in the train and hotel domains. This highlights the versatility of OmniDialog in the task of dialogue summarization across all domains.
    \item[(2)] Among the baselines, Pegasus   \cite{zhang2020pegasus}, DOP   \cite{zhao2022domain}, and ADPL   \cite{zhao2022adpl} outperform OmniDialog in certain domains. All three models possess a significantly larger number of parameters compared to OmniDialog, approximately three times, two times, and three times the parameters of OmniDialog, respectively. Additionally, DOP and ADPL employ domain-specific prompts, whereas OmniDialog maintains consistent prompts across domains.
\end{itemize}

\begin{table*}[htbt]
      \caption{Fine-grained Analysis Results of OmniDialog, OmniDialog-S, OmniDialog-MN, and PPTOD
      on E2E Modeling, DST, Dialogue Summarization (DialogSum and TODSum datasets) tasks. 
      \texttt{``L''} and \texttt{``N''} denote the ``Length'' and ``Number'', respectively.
      }
  \centering \scriptsize
  \renewcommand\tabcolsep{0.4pt}
    \renewcommand\arraystretch{0.80} 
    \resizebox{\textwidth}{!}{%
    \begin{tabular}{cccc cccc cccc cccc cccc cccc cccc cccc cccc cccc cccc cccc cccc cccc}
    \toprule
 \multicolumn{12}{c}{\textbf{E2E Modeling}} & \multicolumn{12}{c}{\textbf{DST}} & \multicolumn{16}{c}{\textbf{DialogSum}} &  \multicolumn{16}{c}{\textbf{TODSum}} \\
  \cmidrule(lr){1-12} \cmidrule(lr){13-24} \cmidrule(lr){25-40}\cmidrule(lr){41-56}
 \multicolumn{4}{c}{\texttt{sp1\_len}} & \multicolumn{4}{c}{\texttt{sp2\_len}} & \multicolumn{4}{c}{\texttt{utr\_num}} &  \multicolumn{4}{c}{\texttt{sp1\_len}} & \multicolumn{4}{c}{\texttt{sp2\_len}} & \multicolumn{4}{c}{\texttt{utr\_num}} &  \multicolumn{4}{c}{\texttt{sp1\_len}} & \multicolumn{4}{c}{\texttt{sp2\_len}} & \multicolumn{4}{c}{\texttt{utr\_num}} &  \multicolumn{4}{c}{\texttt{refe\_len}} & \multicolumn{4}{c}{\texttt{sp1\_len}} & \multicolumn{4}{c}{\texttt{sp2\_len}} & \multicolumn{4}{c}{\texttt{utr\_num}} & \multicolumn{4}{c}{\texttt{refe\_len}} \\
 \cmidrule(lr){1-12} \cmidrule(lr){13-24} \cmidrule(lr){25-40}\cmidrule(lr){41-56}
  \multicolumn{4}{c}{\texttt{L:6-10}} & \multicolumn{4}{c}{\texttt{L:6-10}} & \multicolumn{4}{c}{\texttt{N:2-5}} &  \multicolumn{4}{c}{\texttt{L:6-10}} & \multicolumn{4}{c}{\texttt{L:6-10}} & \multicolumn{4}{c}{\texttt{N:2-5}} &  \multicolumn{4}{c}{\texttt{N:3-30}} & \multicolumn{4}{c}{\texttt{L:1-20}} & \multicolumn{4}{c}{\texttt{N:1-8}} &  \multicolumn{4}{c}{\texttt{L:4-23}} & \multicolumn{4}{c}{\texttt{L:6-9}} & \multicolumn{4}{c}{\texttt{L:6-8}} & \multicolumn{4}{c}{\texttt{N:2-5}} & \multicolumn{4}{c}{\texttt{L:10-34}} \\
 
 \midrule
 \multicolumn{4}{c}{\multirow{5}[2]{*}{\includegraphics[scale=0.067]{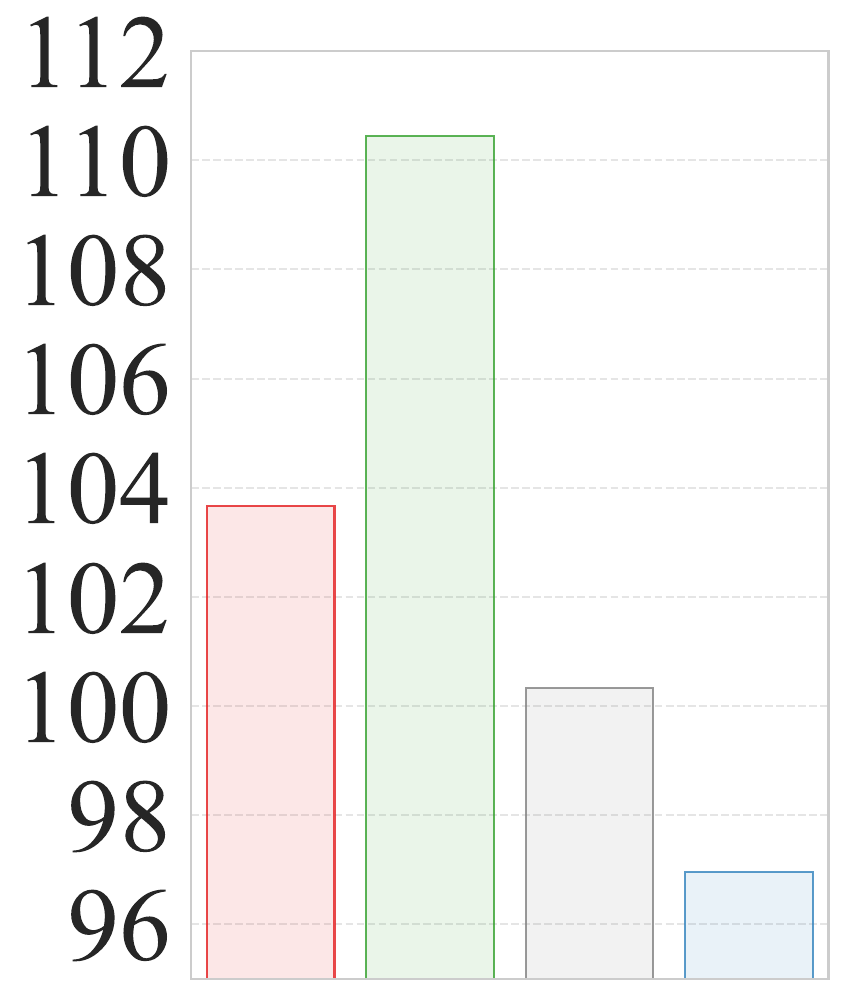}}} & 
\multicolumn{4}{c}{\multirow{5}[2]{*}{\includegraphics[scale=0.067]{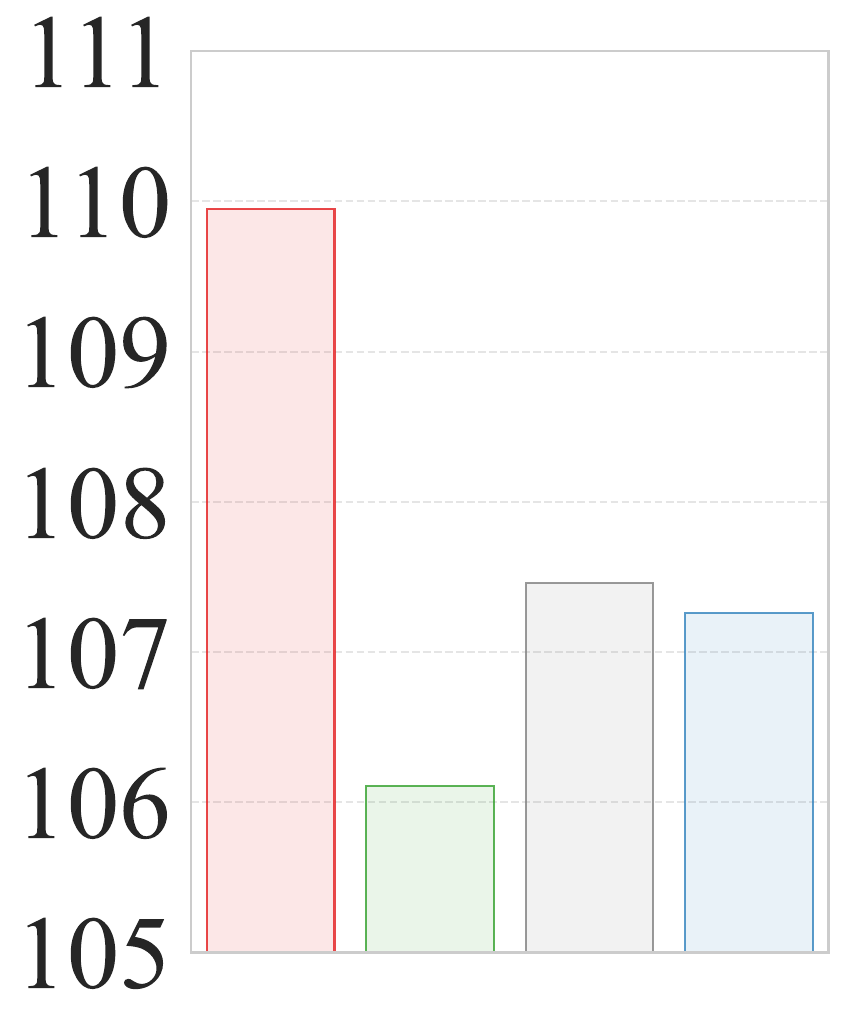}}} &
\multicolumn{4}{c}{\multirow{5}[2]{*}{\includegraphics[scale=0.067]{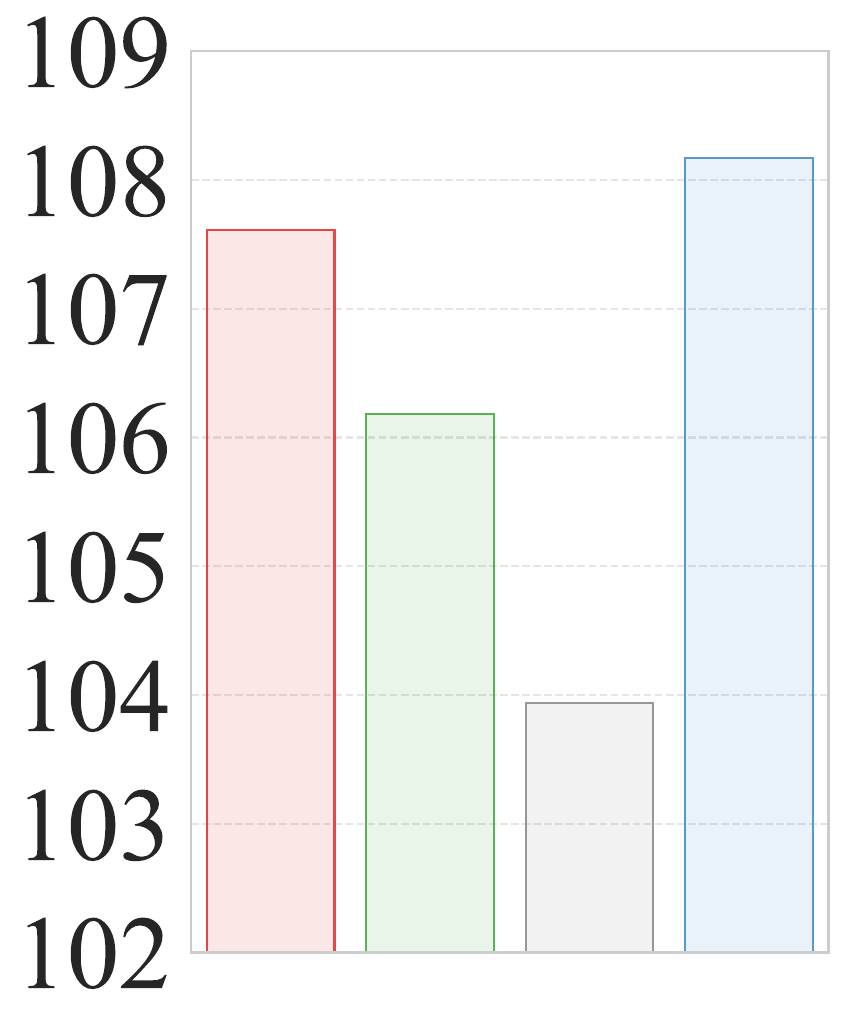}}} & 
 \multicolumn{4}{c}{\multirow{5}[2]{*}{\includegraphics[scale=0.067]{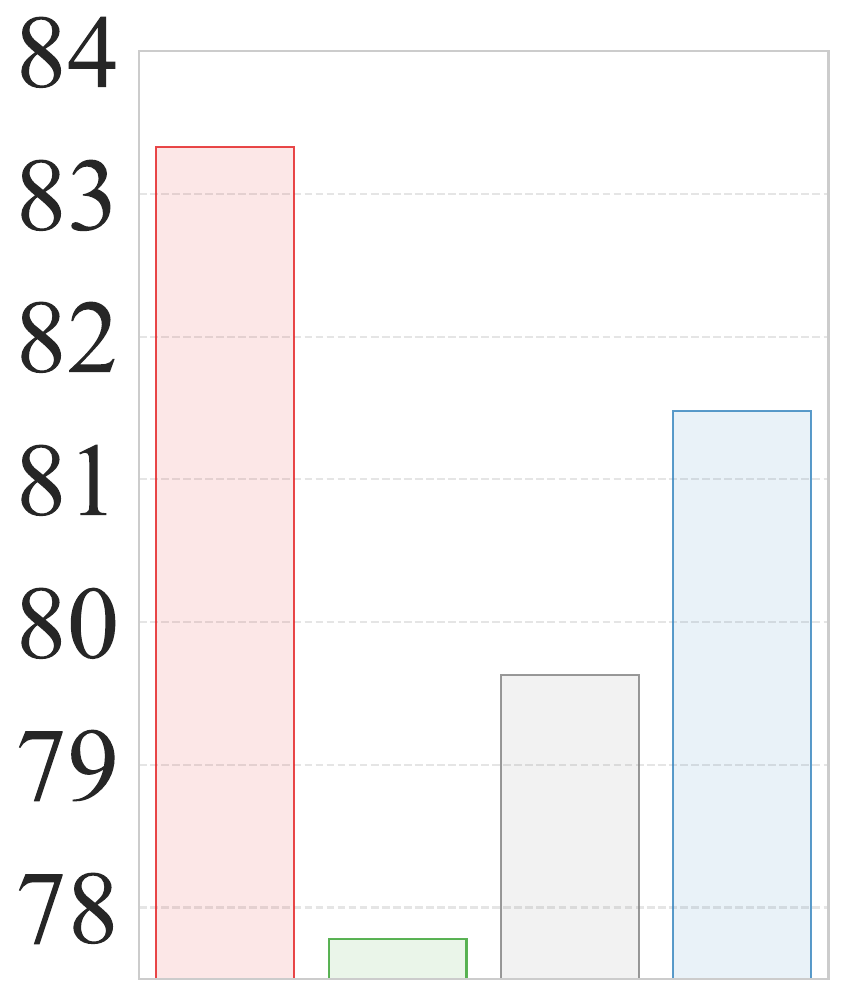}}} & 
\multicolumn{4}{c}{\multirow{5}[2]{*}{\includegraphics[scale=0.067]{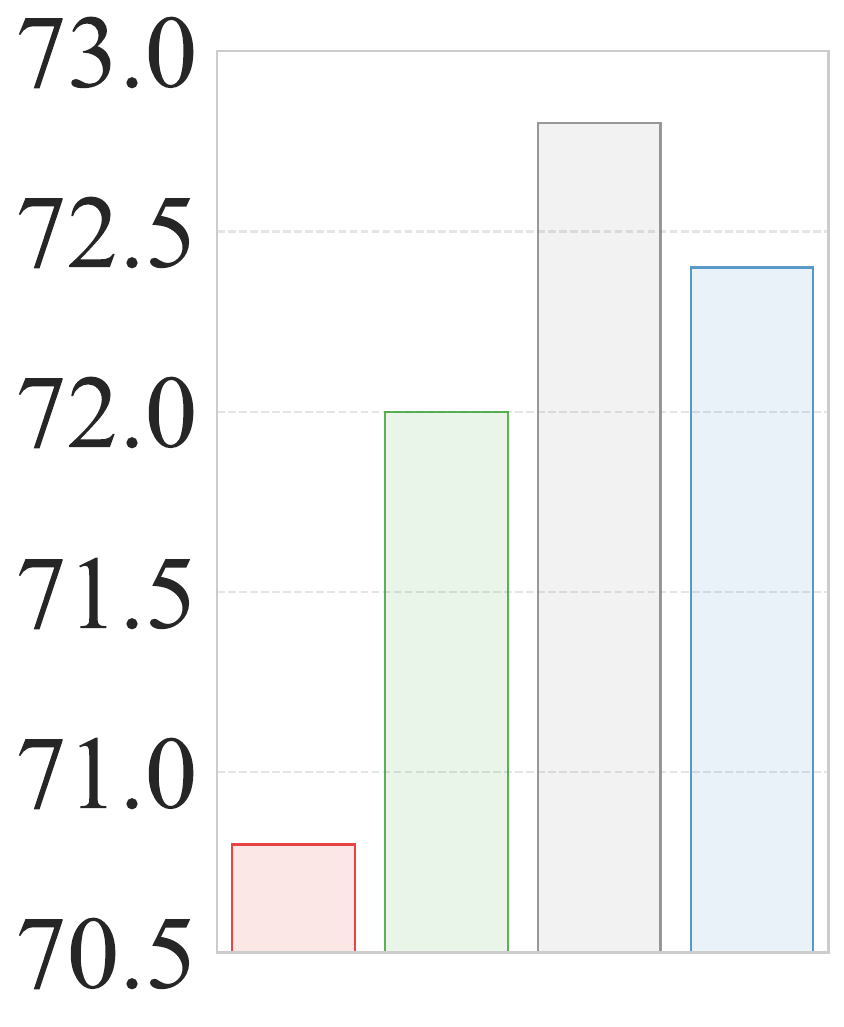}}} &
\multicolumn{4}{c}{\multirow{5}[2]{*}{\includegraphics[scale=0.067]{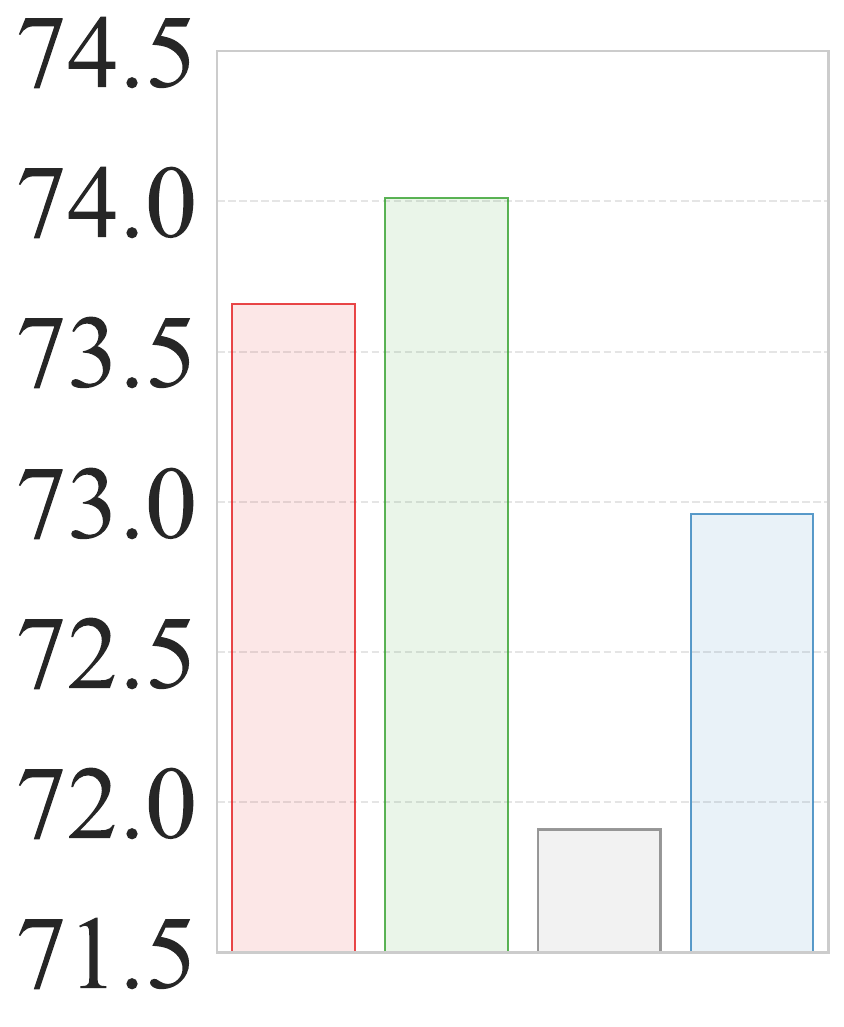}}} &  
\multicolumn{4}{c}{\multirow{5}[2]{*}{\includegraphics[scale=0.067]{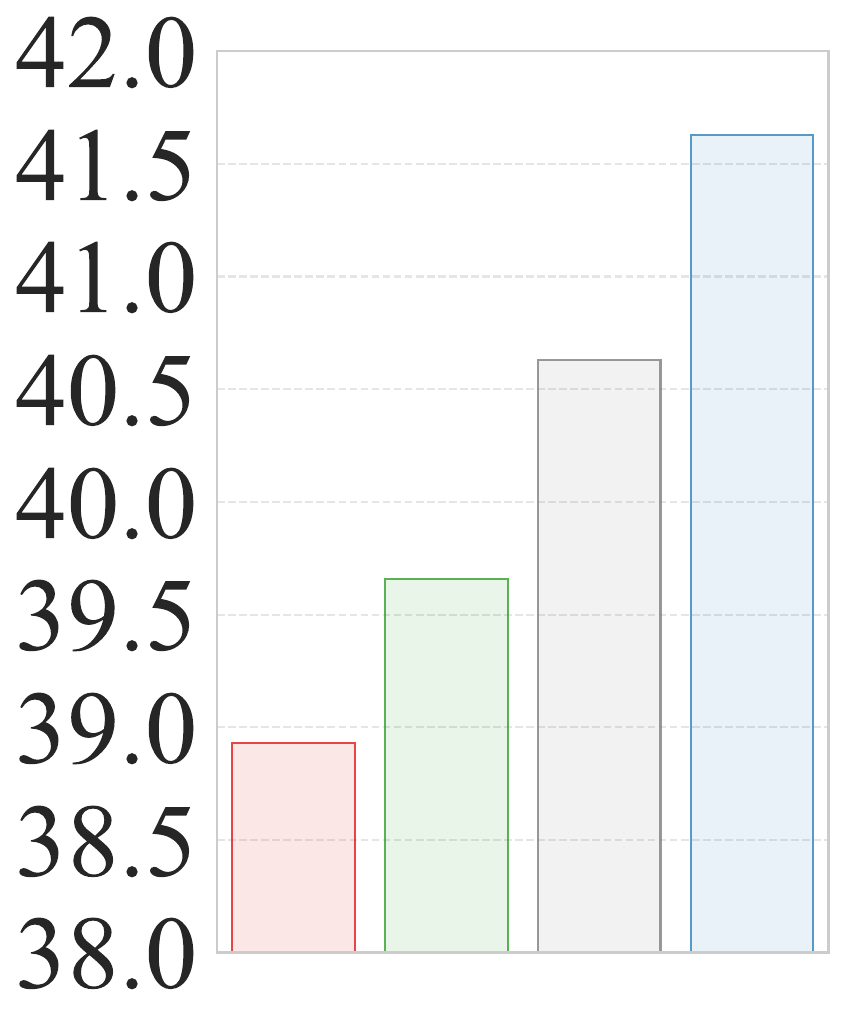}}} &
 \multicolumn{4}{c}{\multirow{5}[2]{*}{\includegraphics[scale=0.067]{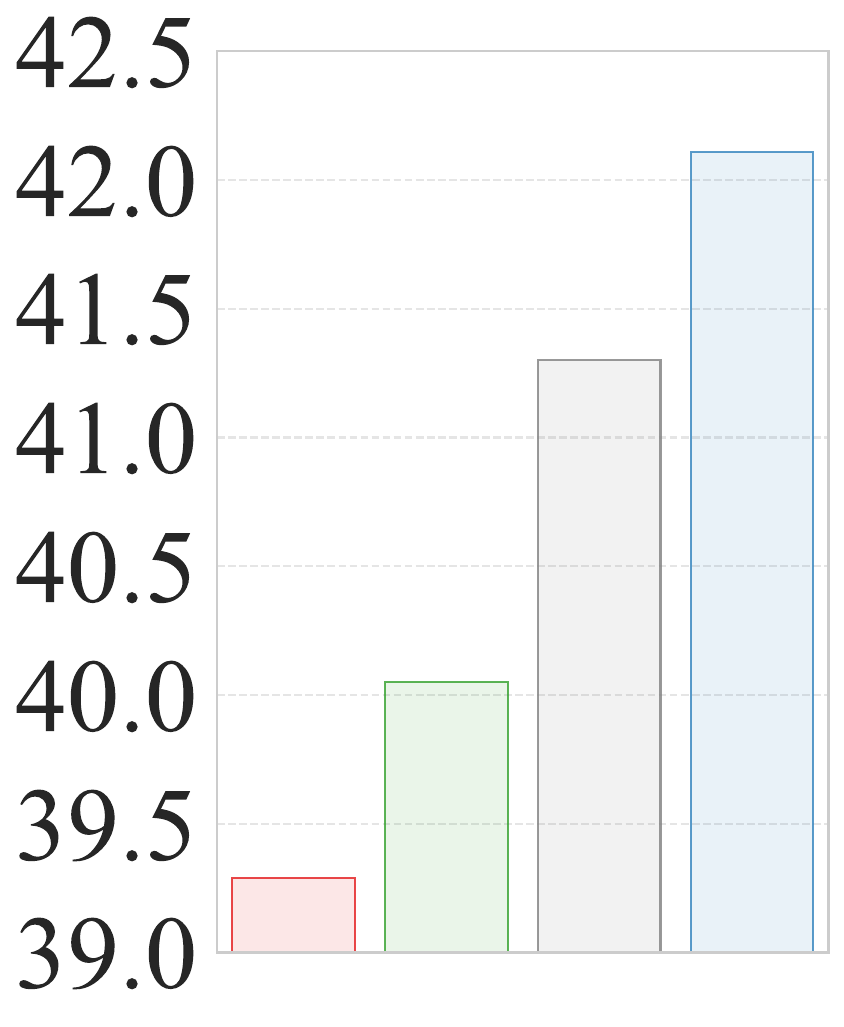}}} & 
\multicolumn{4}{c}{\multirow{5}[2]{*}{\includegraphics[scale=0.067]{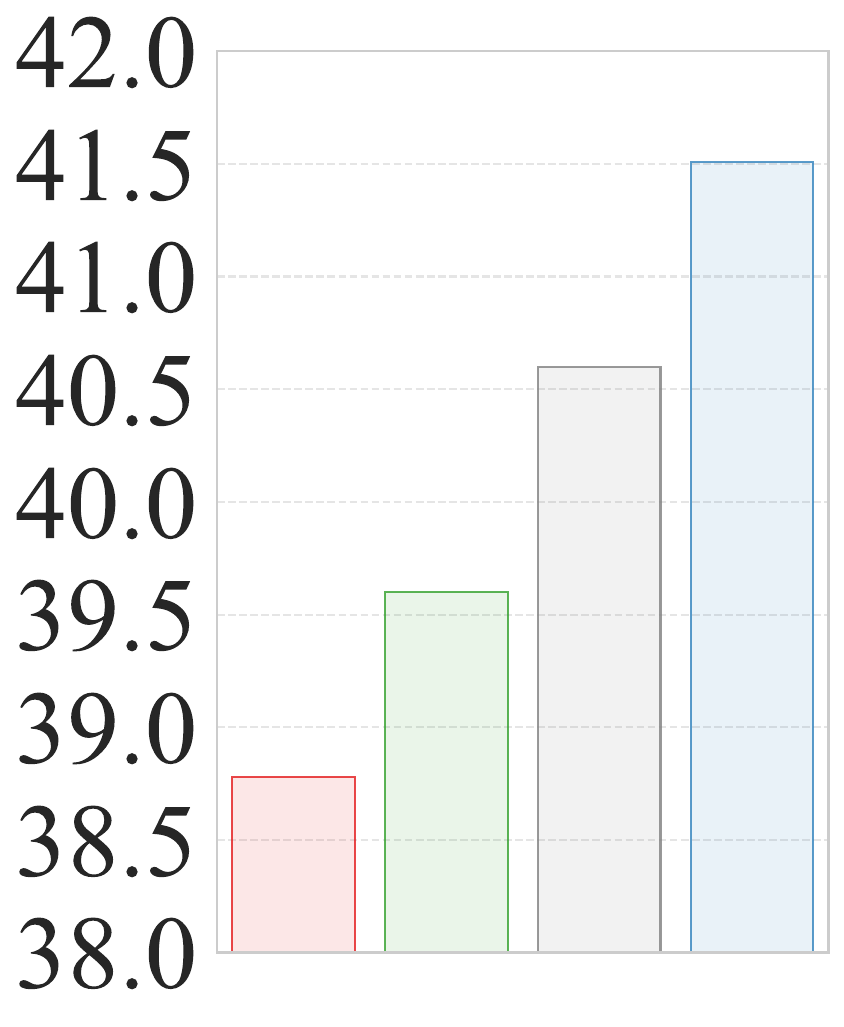}}} &
\multicolumn{4}{c}{\multirow{5}[2]{*}{\includegraphics[scale=0.067]{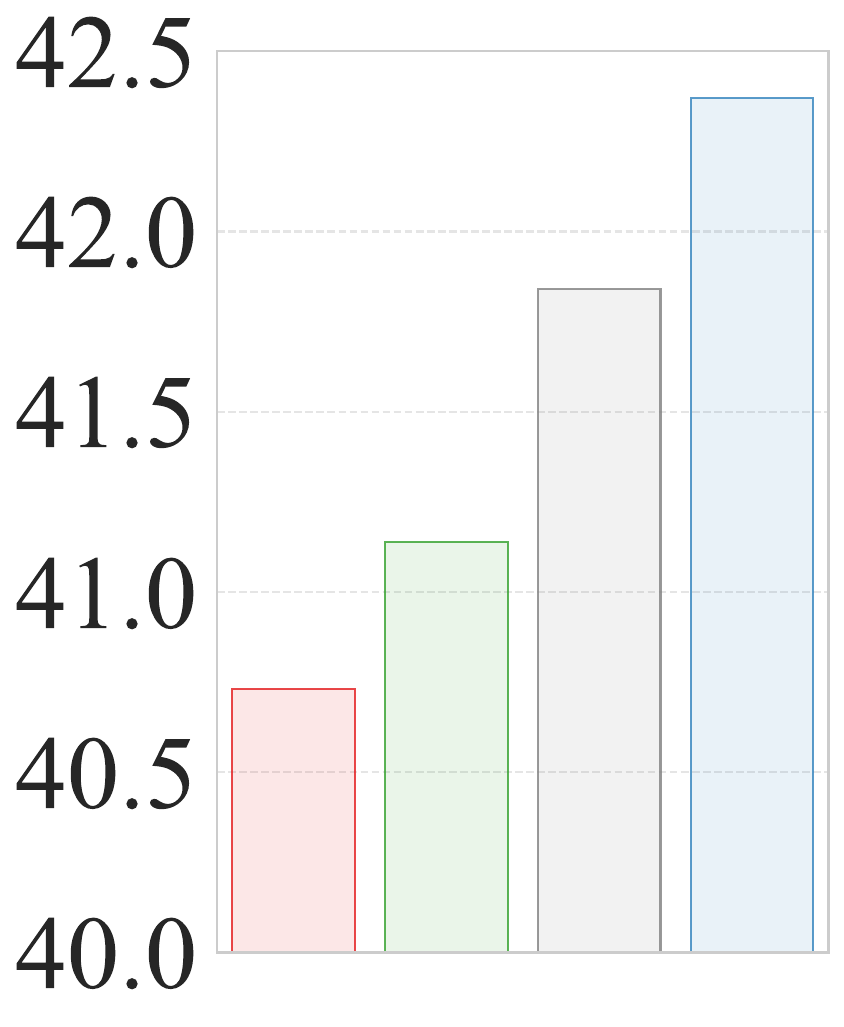}}} & 
 \multicolumn{4}{c}{\multirow{5}[2]{*}{\includegraphics[scale=0.067]{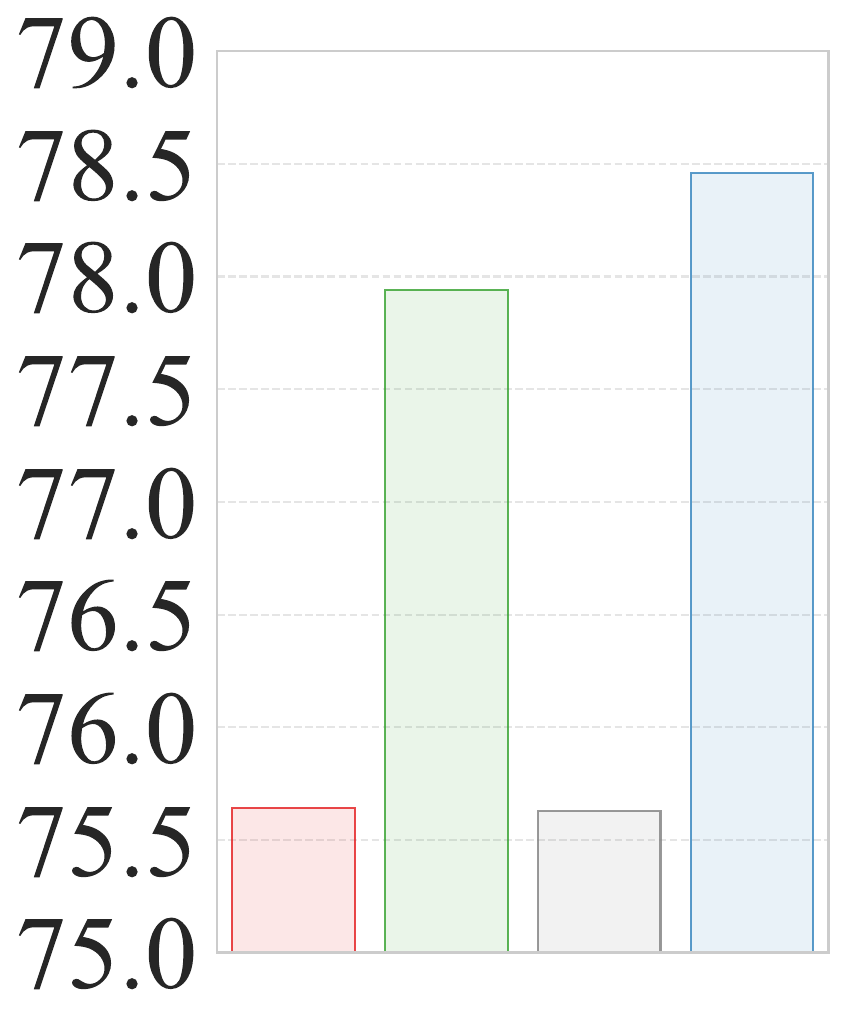}}} & 
\multicolumn{4}{c}{\multirow{5}[2]{*}{\includegraphics[scale=0.067]{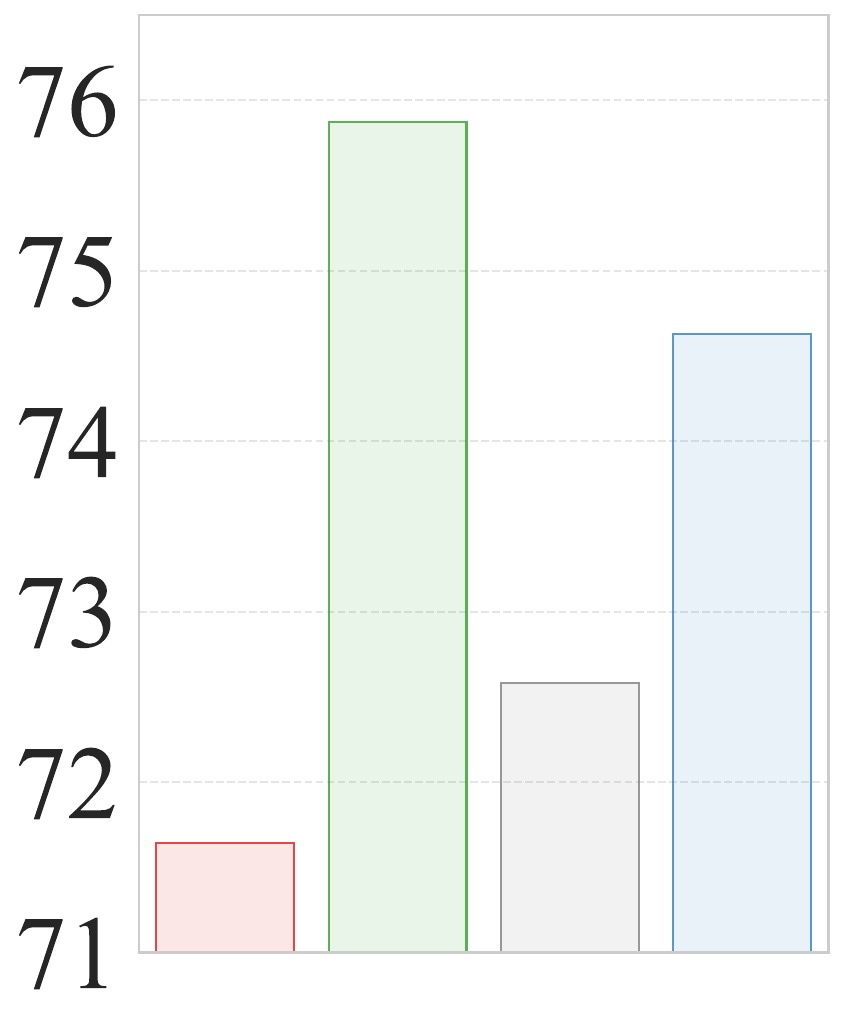}}} &
\multicolumn{4}{c}{\multirow{5}[2]{*}{\includegraphics[scale=0.067]{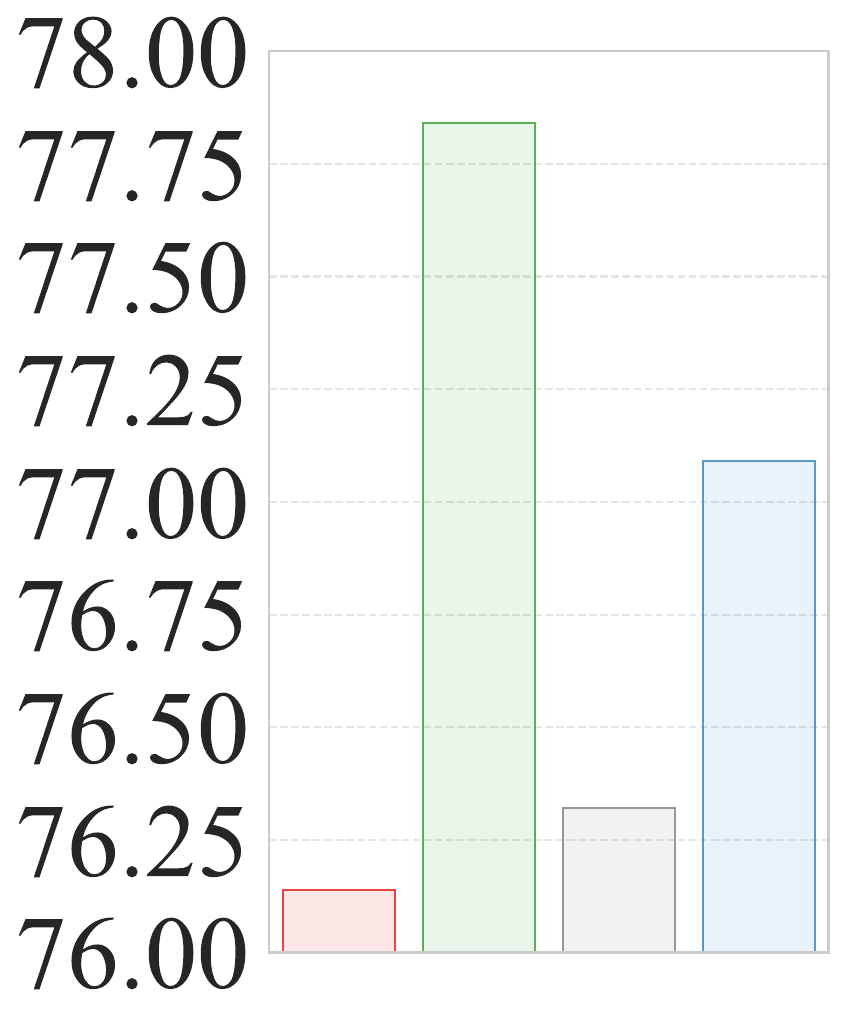}}} &  
\multicolumn{4}{c}{\multirow{5}[2]{*}{\includegraphics[scale=0.067]{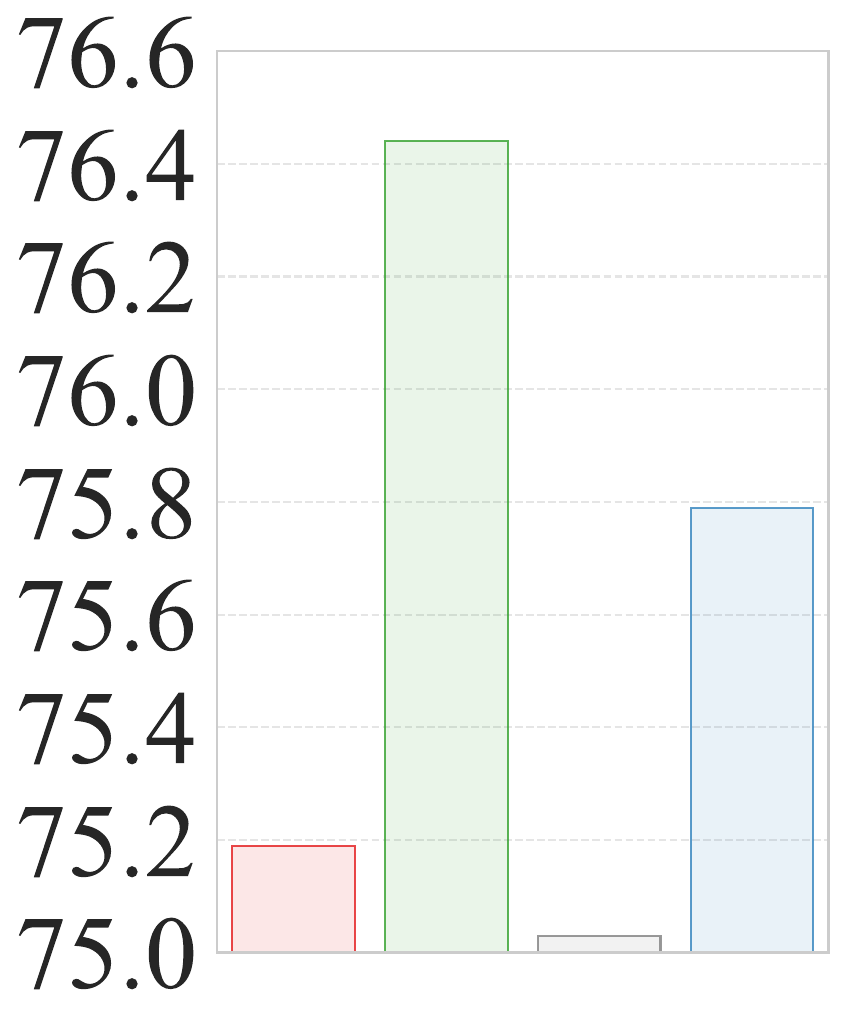}}} 
\\ \\ \\ \\ \\ \\ \\
    \midrule
         
 \multicolumn{4}{c}{\texttt{L:11-15}} & \multicolumn{4}{c}{\texttt{L:11-15}} & \multicolumn{4}{c}{\texttt{N:6-9}} &  \multicolumn{4}{c}{\texttt{L:11-15}} & \multicolumn{4}{c}{\texttt{L:11-15}} & \multicolumn{4}{c}{\texttt{N:6-9}} &  \multicolumn{4}{c}{\texttt{L:31-60}} & \multicolumn{4}{c}{\texttt{L:21-40}} & \multicolumn{4}{c}{\texttt{N:9-16}} &  \multicolumn{4}{c}{\texttt{L:24-43}} & \multicolumn{4}{c}{\texttt{L:10-13}} & \multicolumn{4}{c}{\texttt{L:9-11}} & \multicolumn{4}{c}{\texttt{N:6-9}} & \multicolumn{4}{c}{\texttt{L:35-59}} \\
 \midrule

    % \midrule
        \multicolumn{4}{c}{\multirow{5}[2]{*}{\includegraphics[scale=0.067]{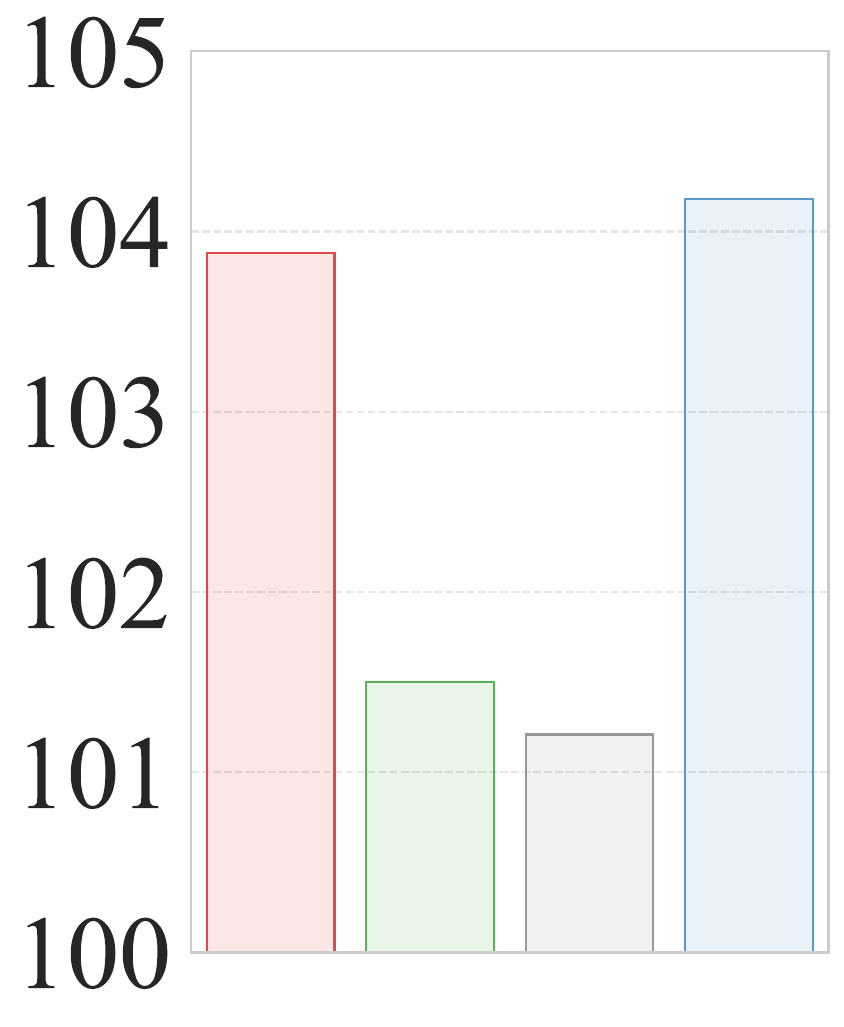}}} & 
        \multicolumn{4}{c}{\multirow{5}[2]{*}{\includegraphics[scale=0.067]{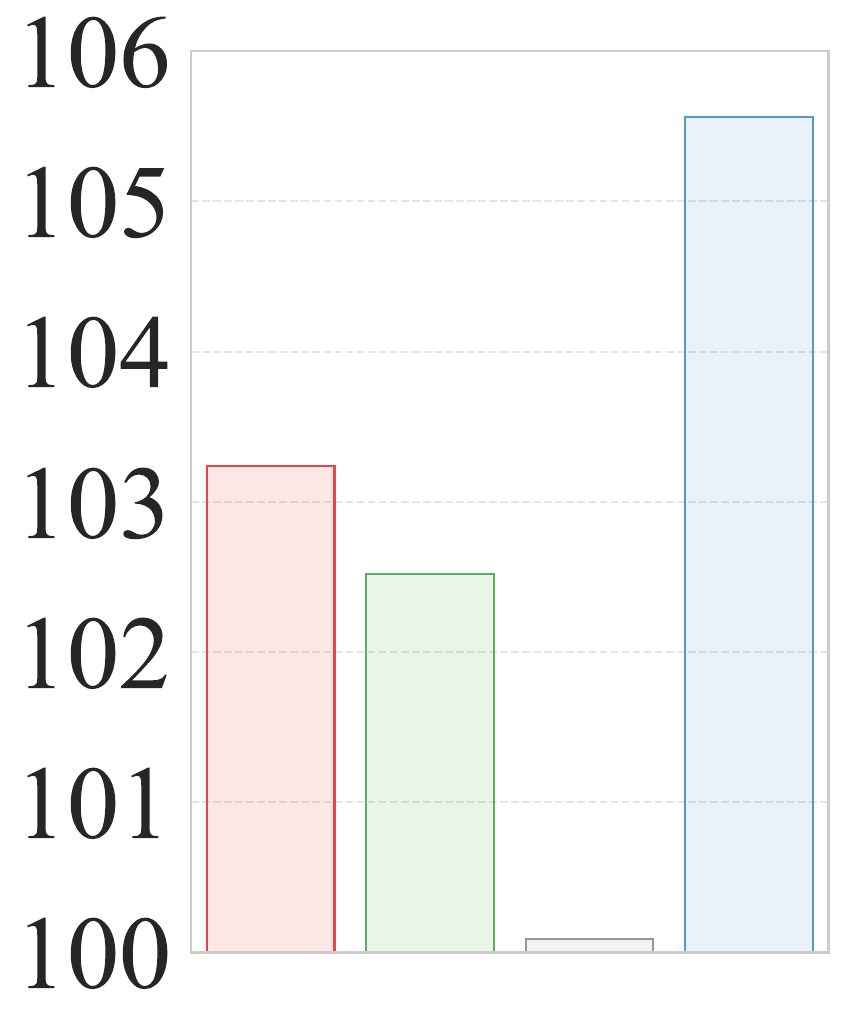}}} &
         \multicolumn{4}{c}{\multirow{5}[2]{*}{\includegraphics[scale=0.067]{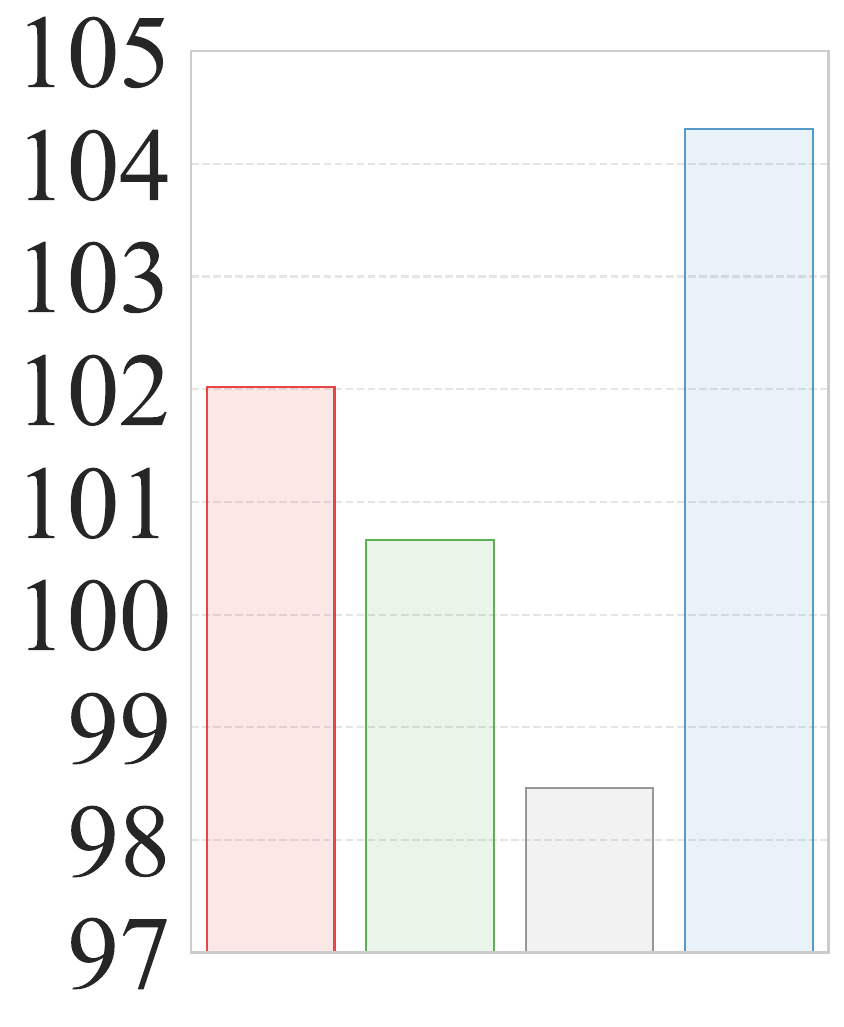}}} & 
                 \multicolumn{4}{c}{\multirow{5}[2]{*}{\includegraphics[scale=0.067]{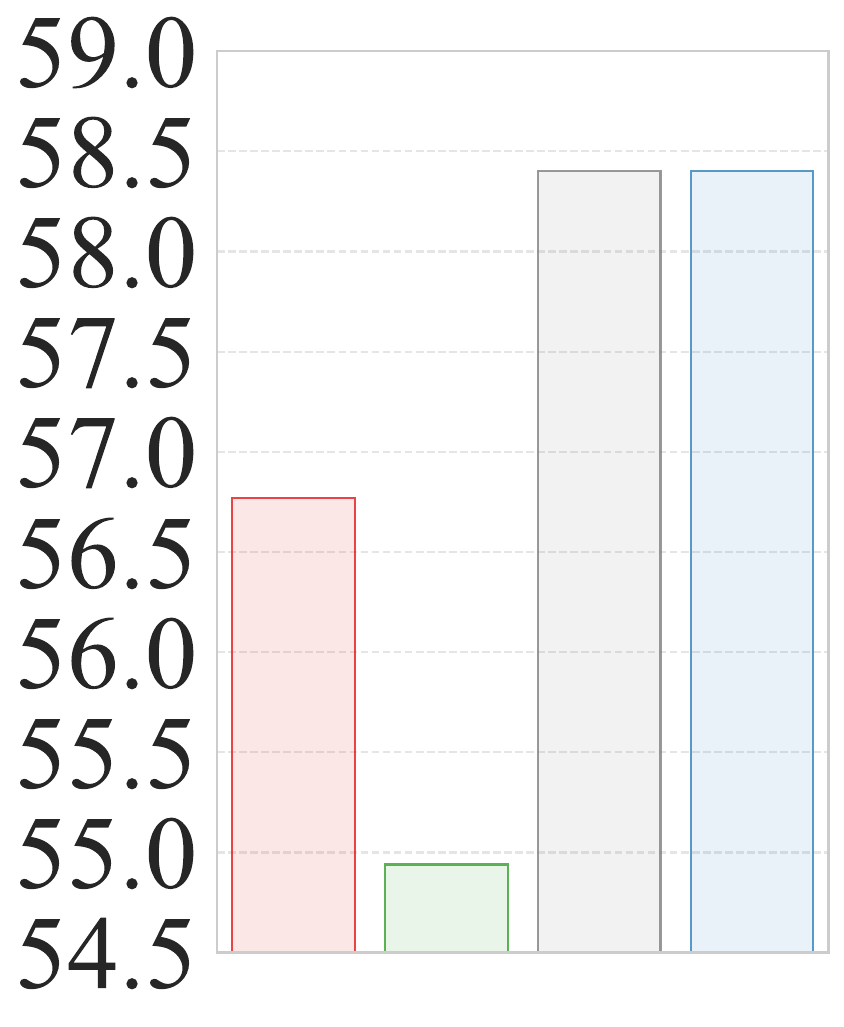}}} & 
        \multicolumn{4}{c}{\multirow{5}[2]{*}{\includegraphics[scale=0.067]{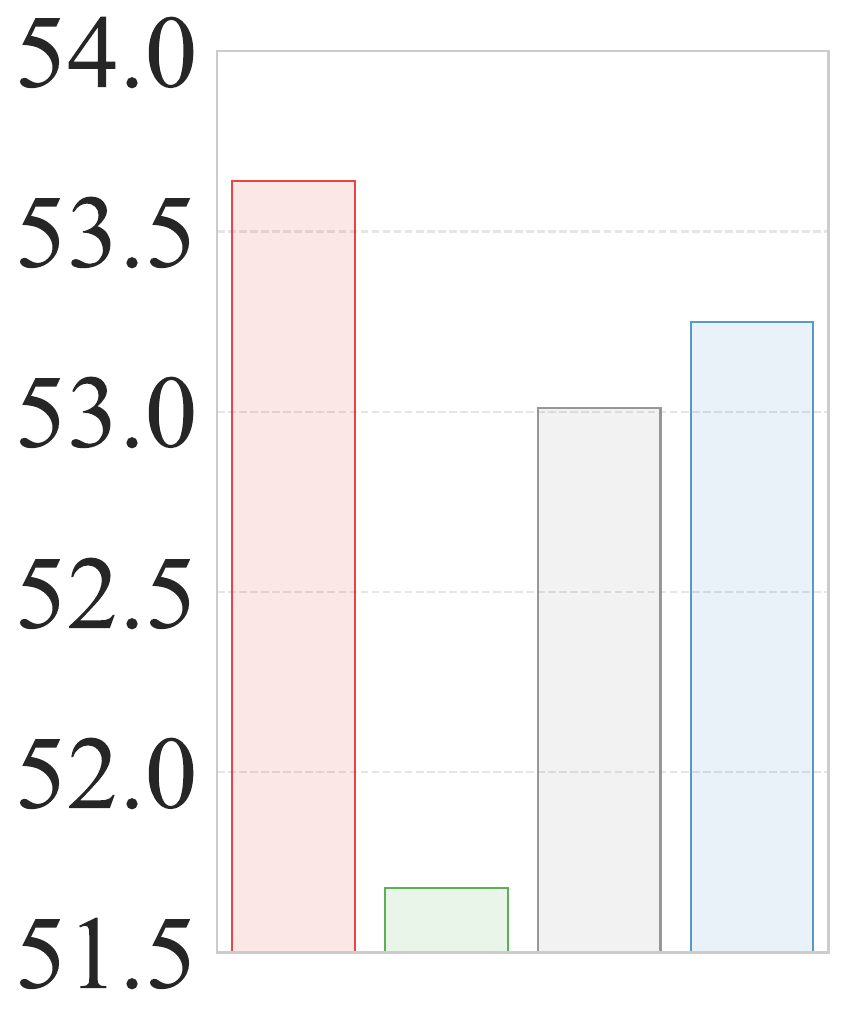}}} &
         \multicolumn{4}{c}{\multirow{5}[2]{*}{\includegraphics[scale=0.067]{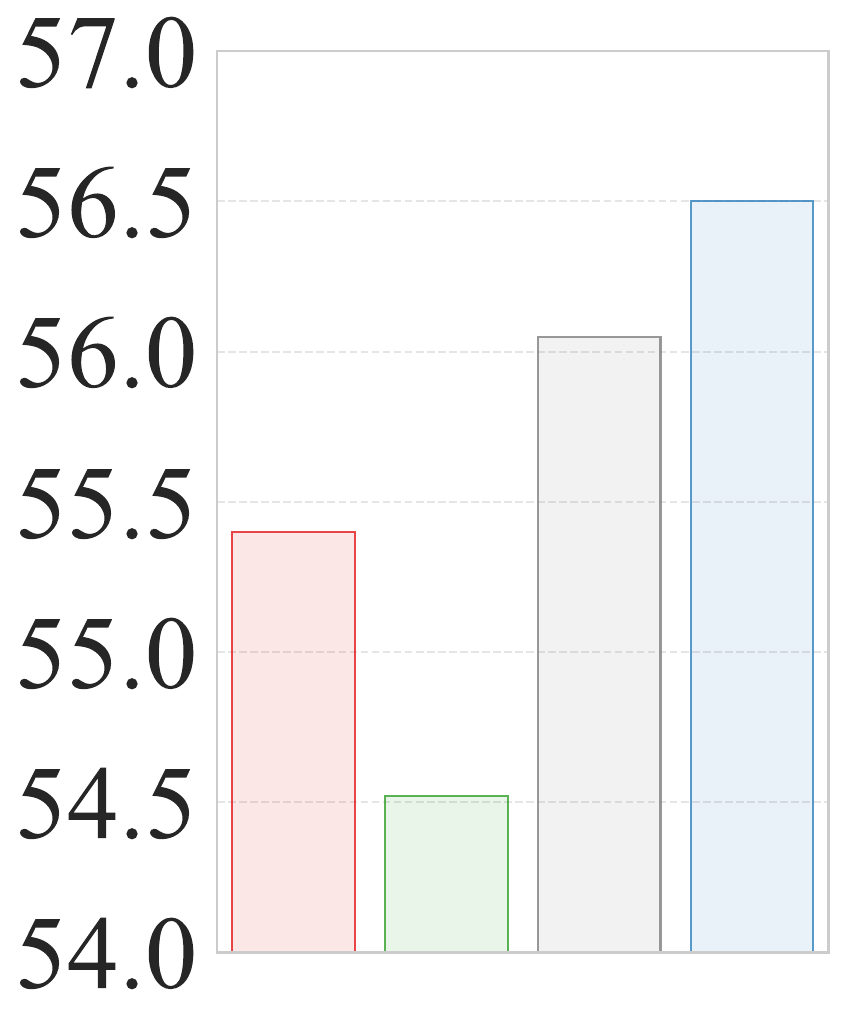}}} & 
         \multicolumn{4}{c}{\multirow{5}[2]{*}{\includegraphics[scale=0.067]{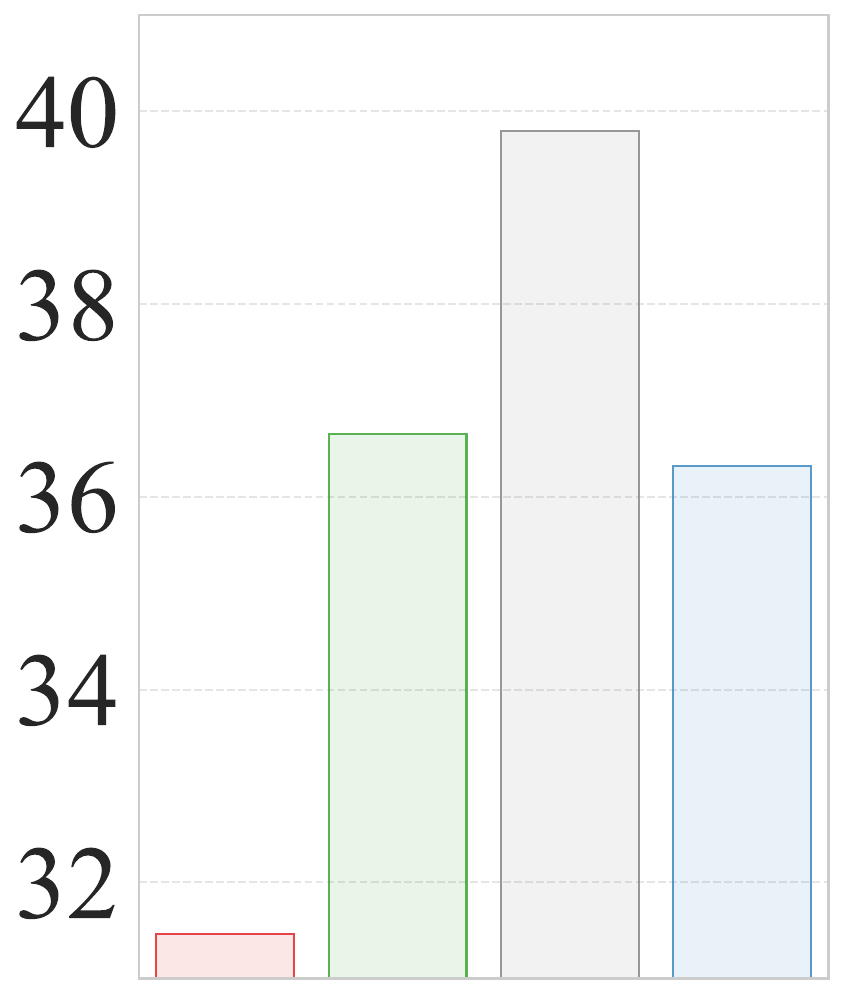}}} &
         \multicolumn{4}{c}{\multirow{5}[2]{*}{\includegraphics[scale=0.067]{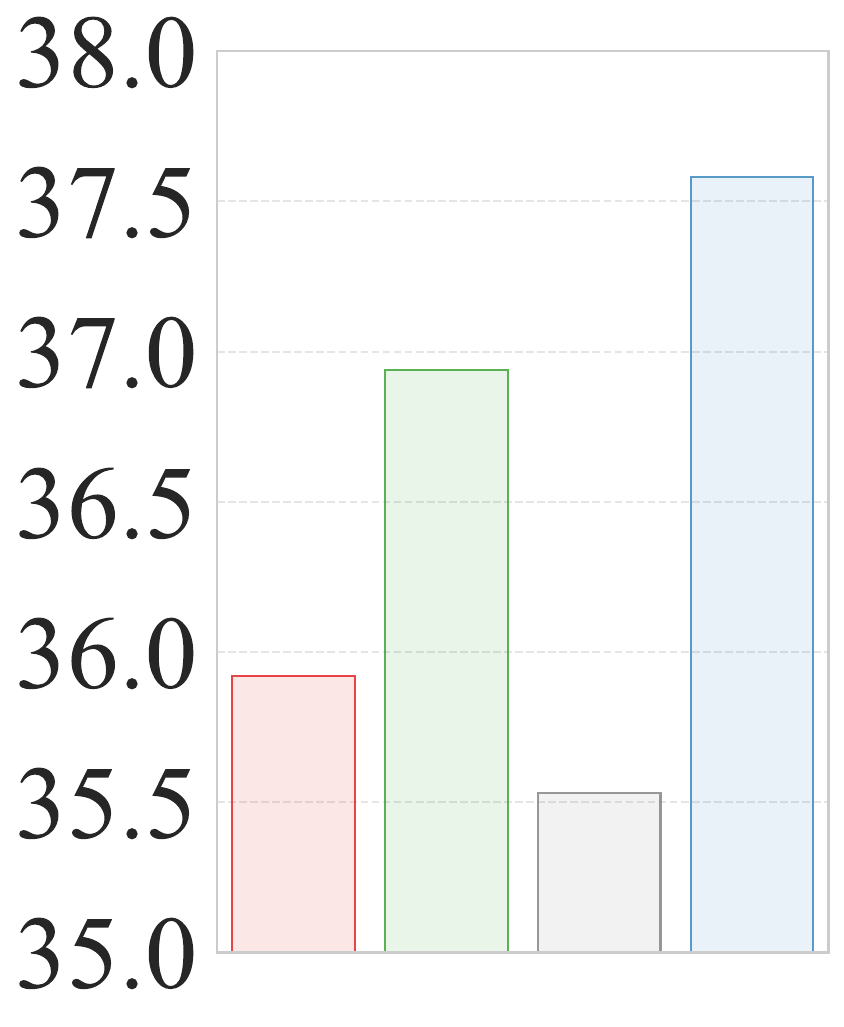}}} & 
        \multicolumn{4}{c}{\multirow{5}[2]{*}{\includegraphics[scale=0.067]{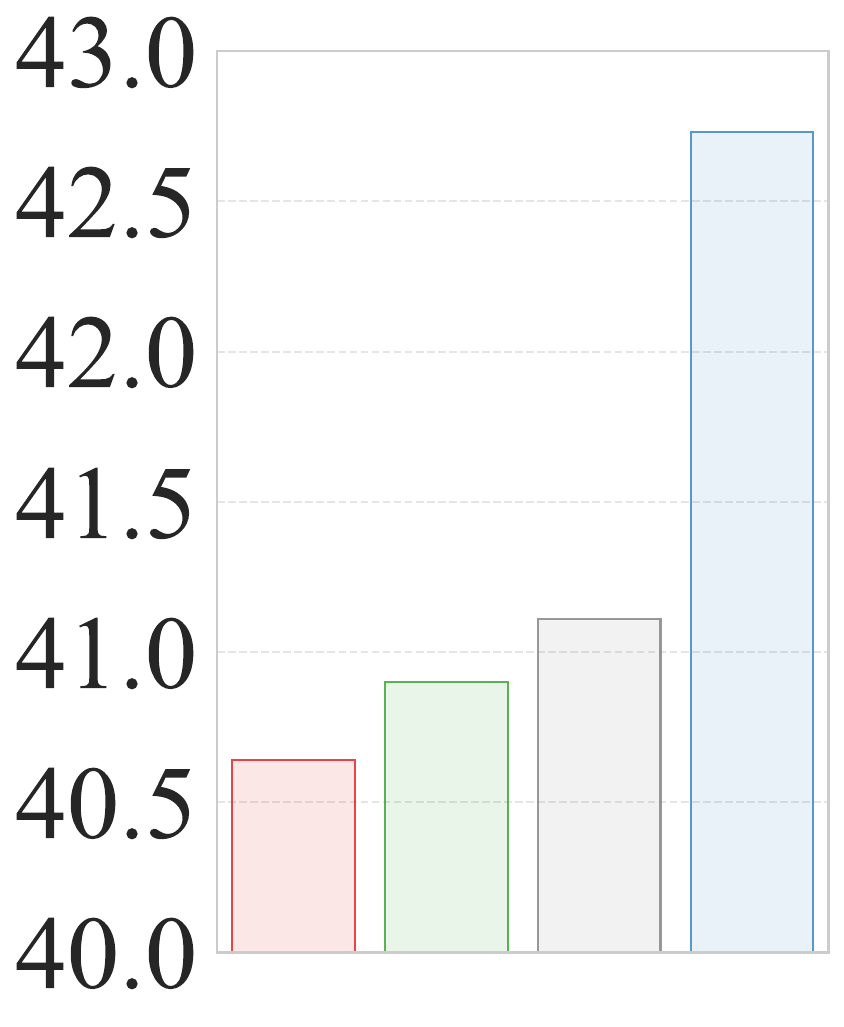}}} &
         \multicolumn{4}{c}{\multirow{5}[2]{*}{\includegraphics[scale=0.067]{pic/dialogsum/reference/24-43.pdf}}} & 
                 \multicolumn{4}{c}{\multirow{5}[2]{*}{\includegraphics[scale=0.067]{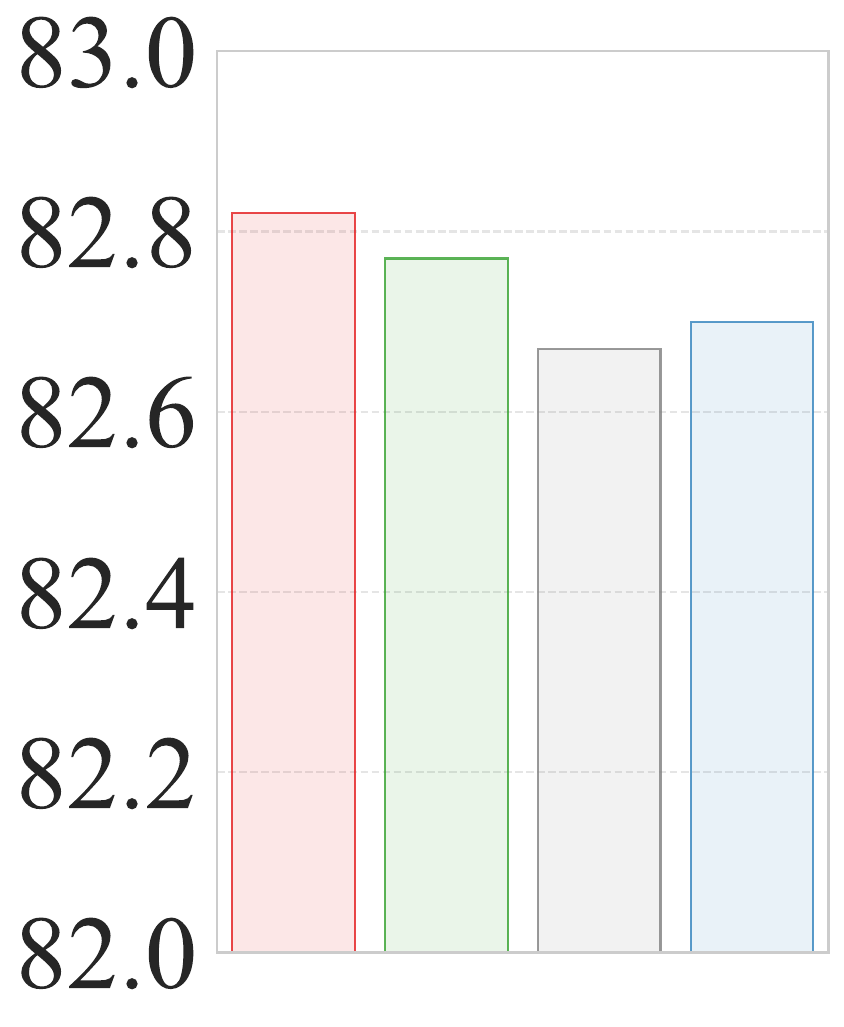}}} & 
        \multicolumn{4}{c}{\multirow{5}[2]{*}{\includegraphics[scale=0.067]{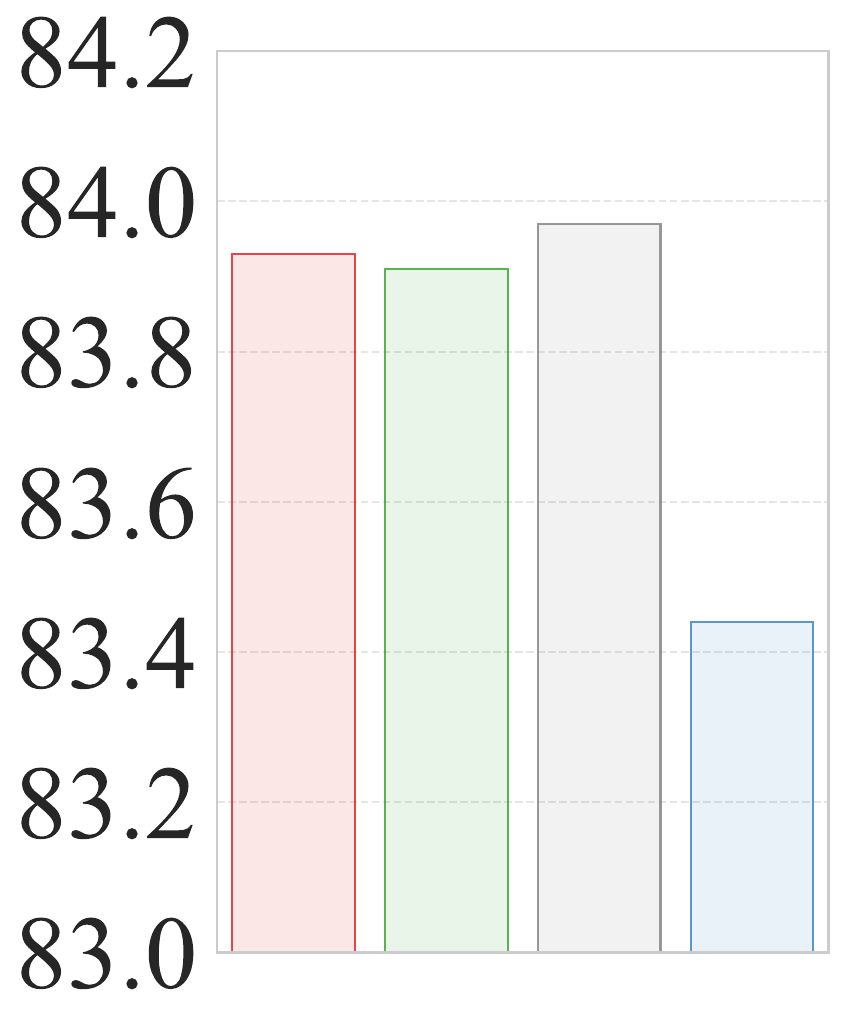}}} &
         \multicolumn{4}{c}{\multirow{5}[2]{*}{\includegraphics[scale=0.067]{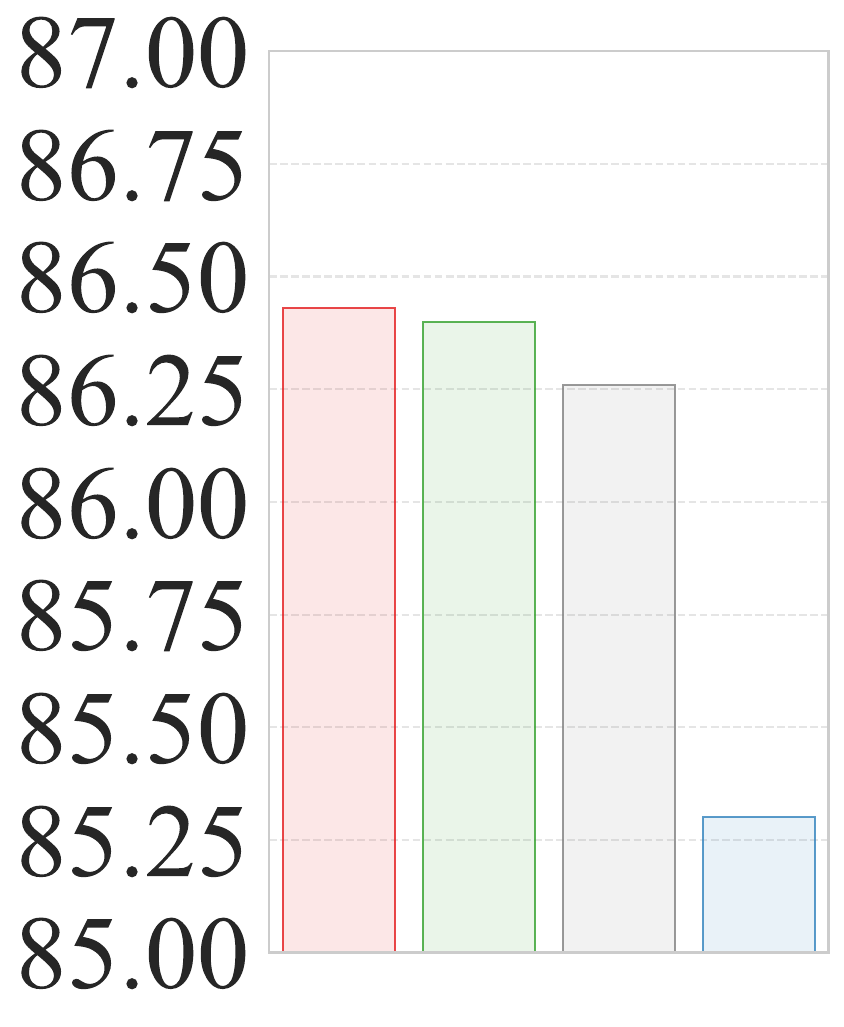}}} & 
         \multicolumn{4}{c}{\multirow{5}[2]{*}{\includegraphics[scale=0.067]{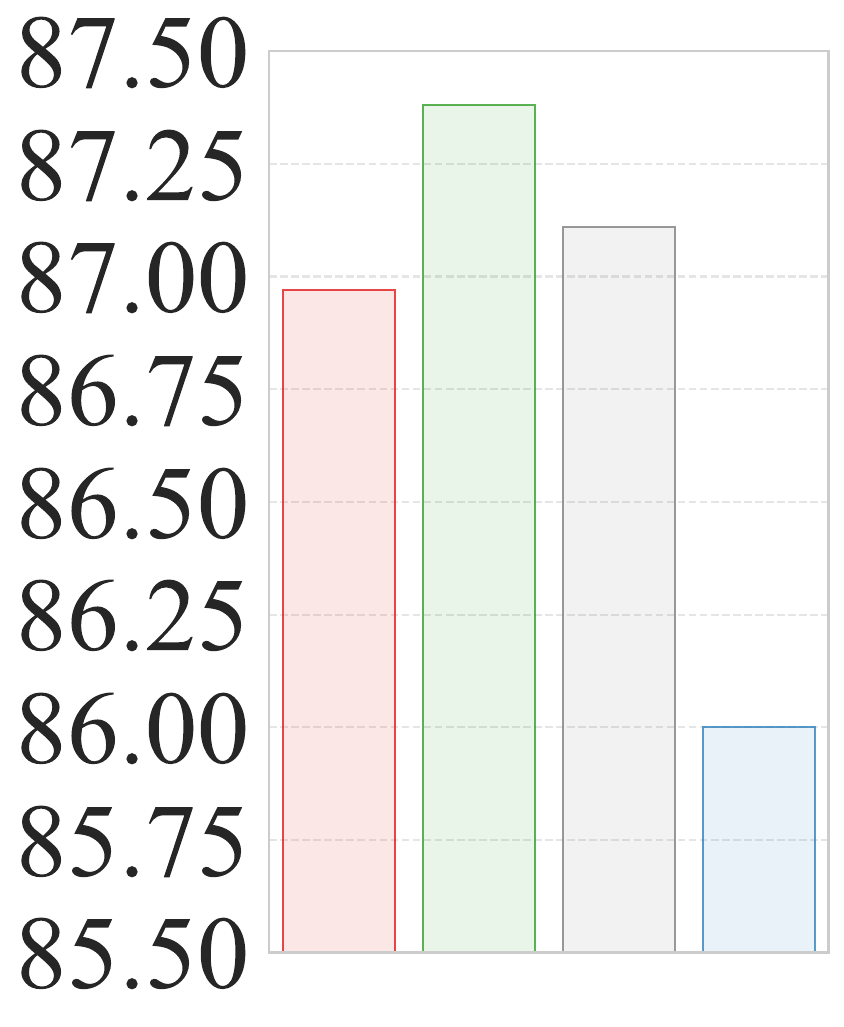}}} 
 \\ \\ \\ \\ \\ \\
 
\midrule

 \multicolumn{4}{c}{\texttt{L:16+}} & \multicolumn{4}{c}{\texttt{L:16+}} & \multicolumn{4}{c}{\texttt{N:10+}} &  \multicolumn{4}{c}{\texttt{L:16+}} & \multicolumn{4}{c}{\texttt{L:16+}} & \multicolumn{4}{c}{\texttt{N:10+}} &  \multicolumn{4}{c}{\texttt{L:61+}} & \multicolumn{4}{c}{\texttt{L:41+}} & \multicolumn{4}{c}{\texttt{N:17+}} &  \multicolumn{4}{c}{\texttt{L:44+}} & \multicolumn{4}{c}{\texttt{L:14+}} & \multicolumn{4}{c}{\texttt{L:12+}} & \multicolumn{4}{c}{\texttt{N:10+}} & \multicolumn{4}{c}{\texttt{L:60+}} \\
\midrule
\multicolumn{4}{c}{\multirow{5}[2]{*}{\includegraphics[scale=0.067]{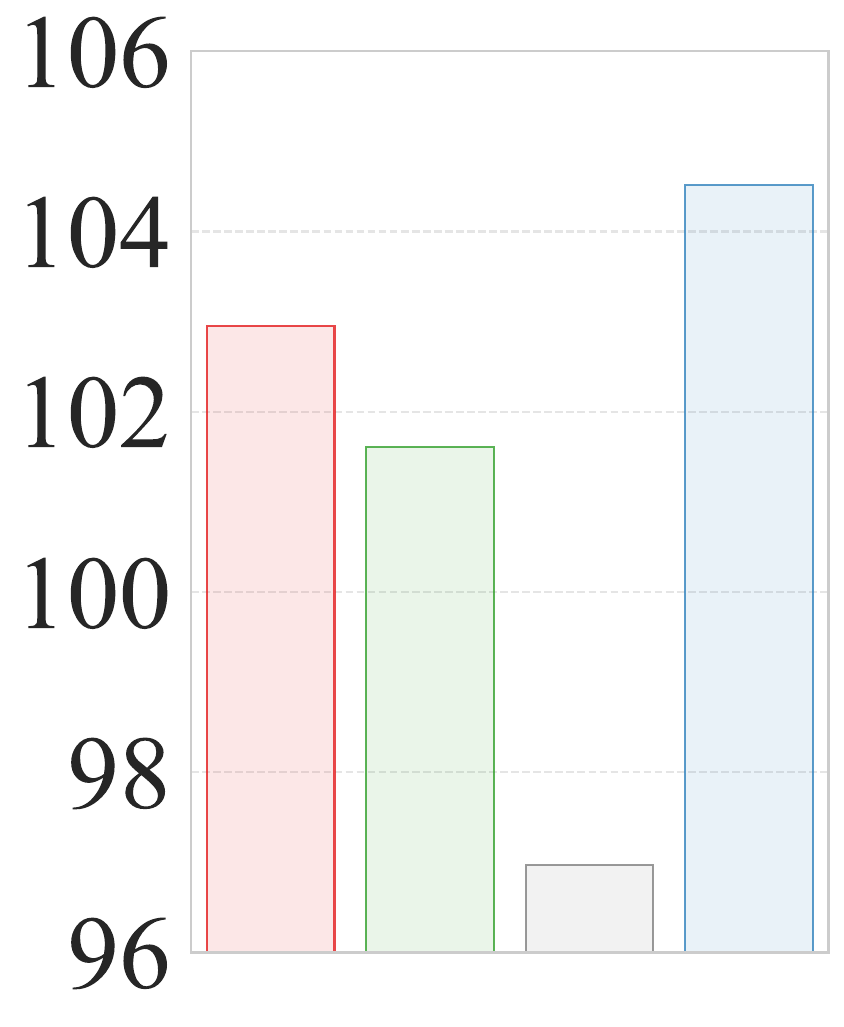}}} & 
\multicolumn{4}{c}{\multirow{5}[2]{*}{\includegraphics[scale=0.067]{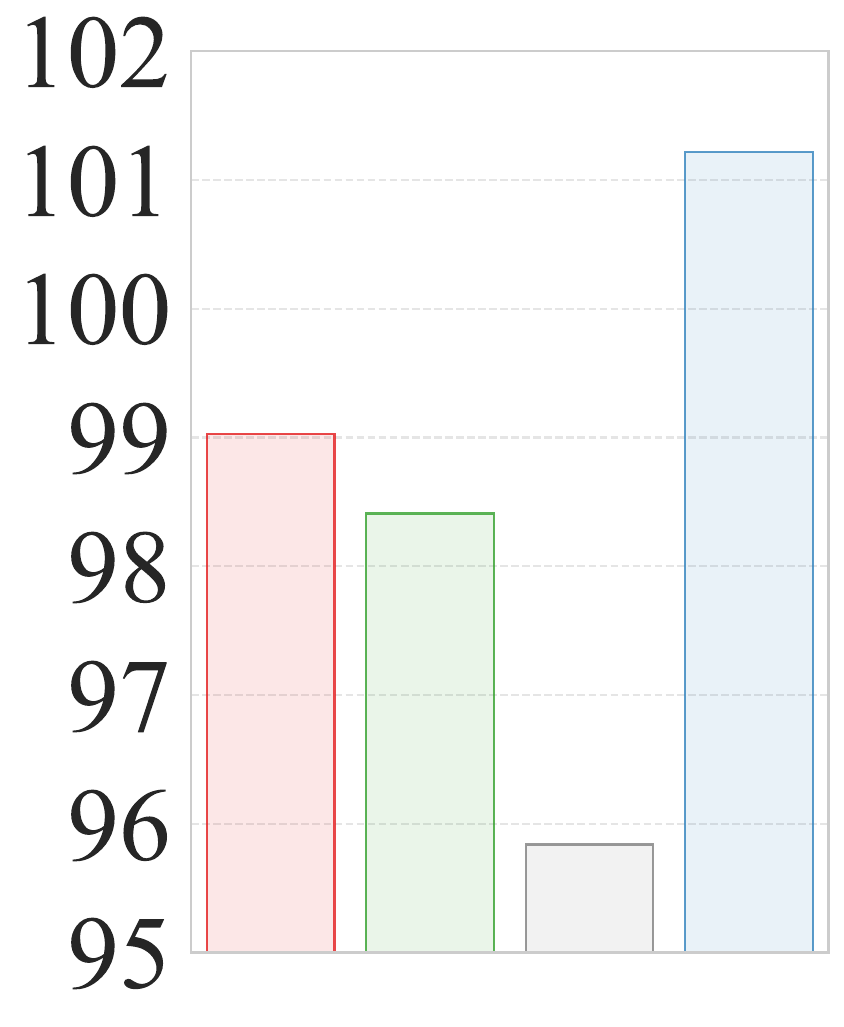}}} &
         \multicolumn{4}{c}{\multirow{5}[2]{*}{\includegraphics[scale=0.067]{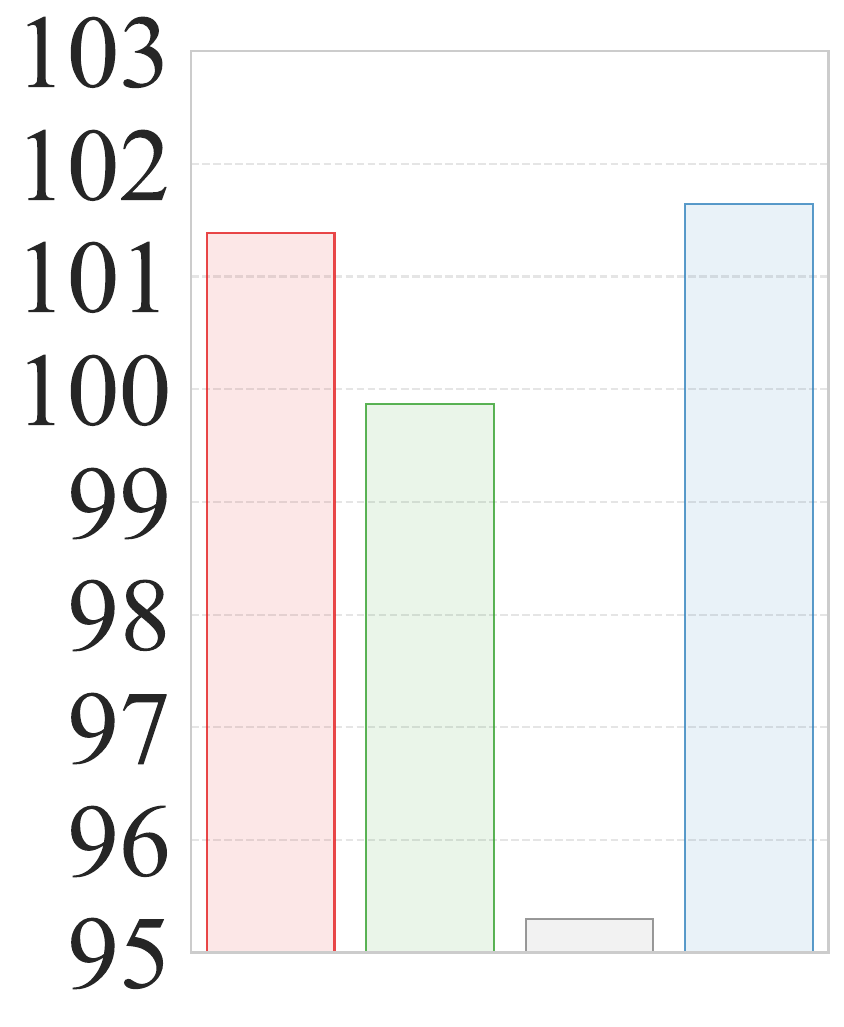}}} & 
         \multicolumn{4}{c}{\multirow{5}[2]{*}{\includegraphics[scale=0.067]{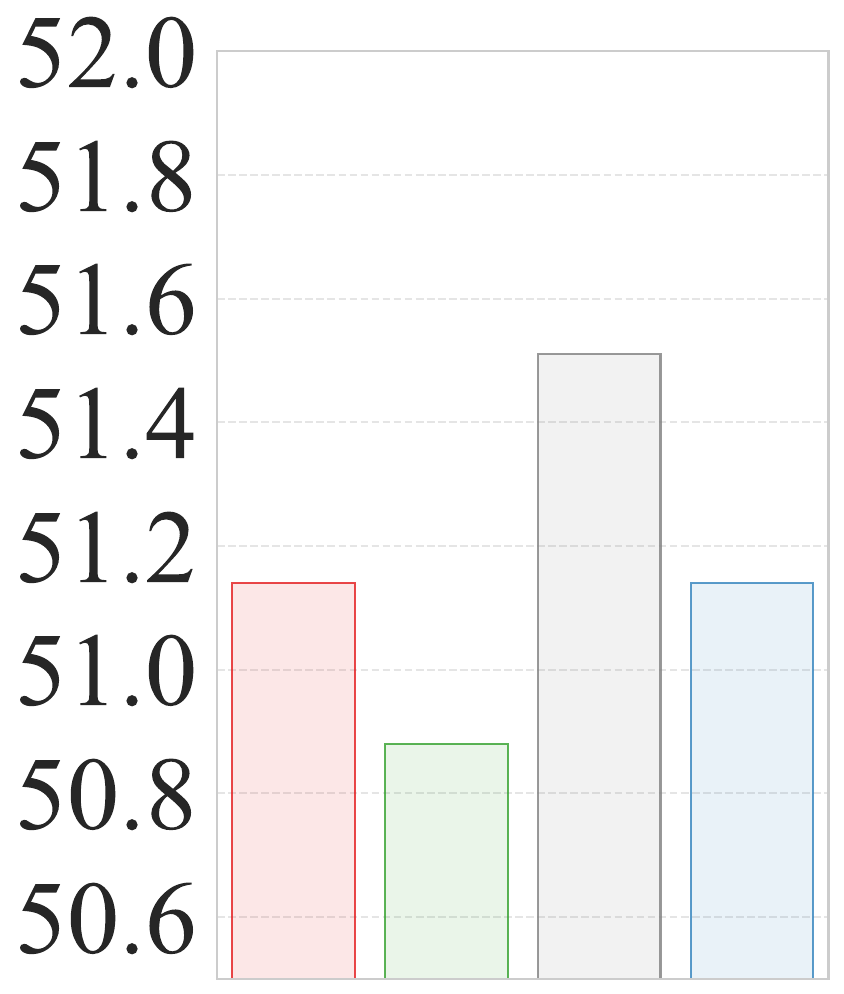}}} & 
        \multicolumn{4}{c}{\multirow{5}[2]{*}{\includegraphics[scale=0.067]{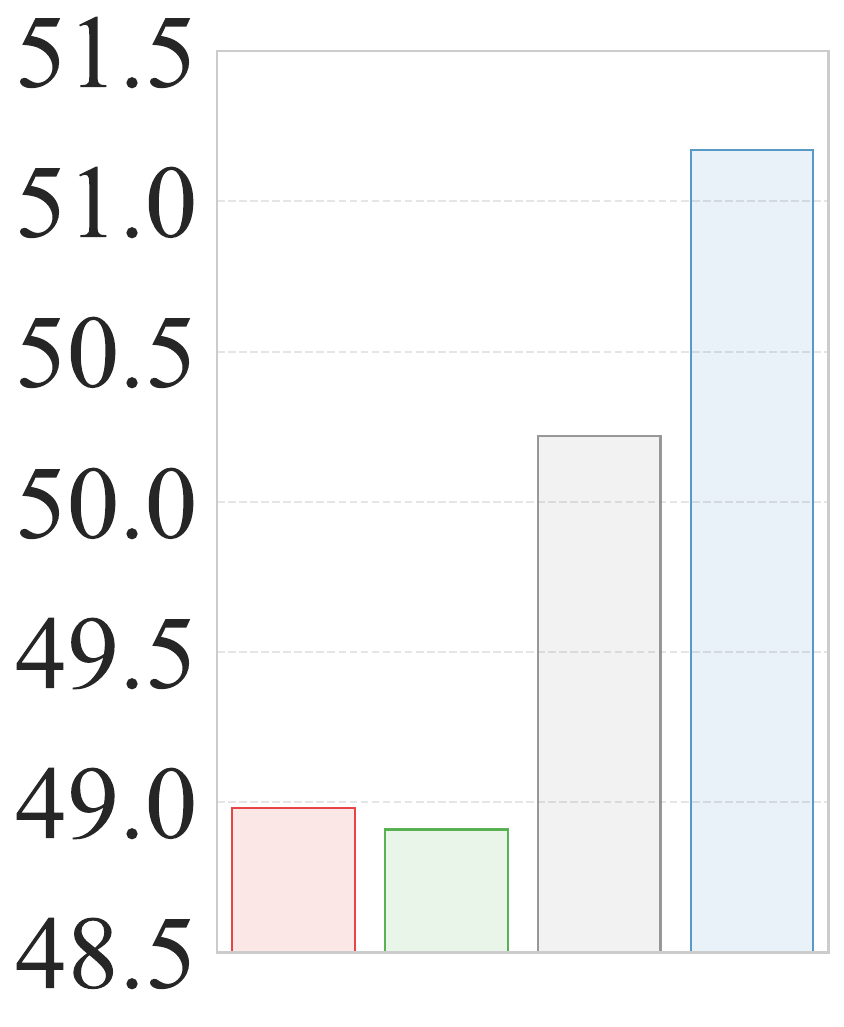}}} &
         \multicolumn{4}{c}{\multirow{5}[2]{*}{\includegraphics[scale=0.067]{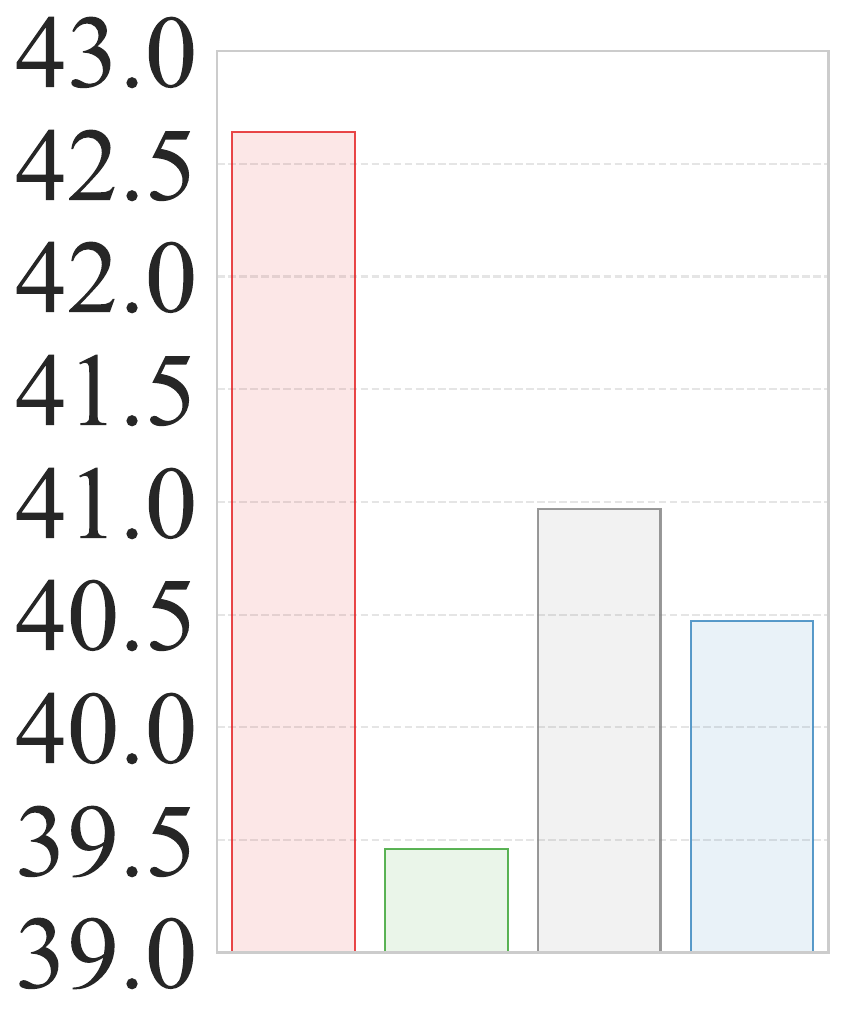}}} & 
         
         \multicolumn{4}{c}{\multirow{5}[2]{*}{\includegraphics[scale=0.067]{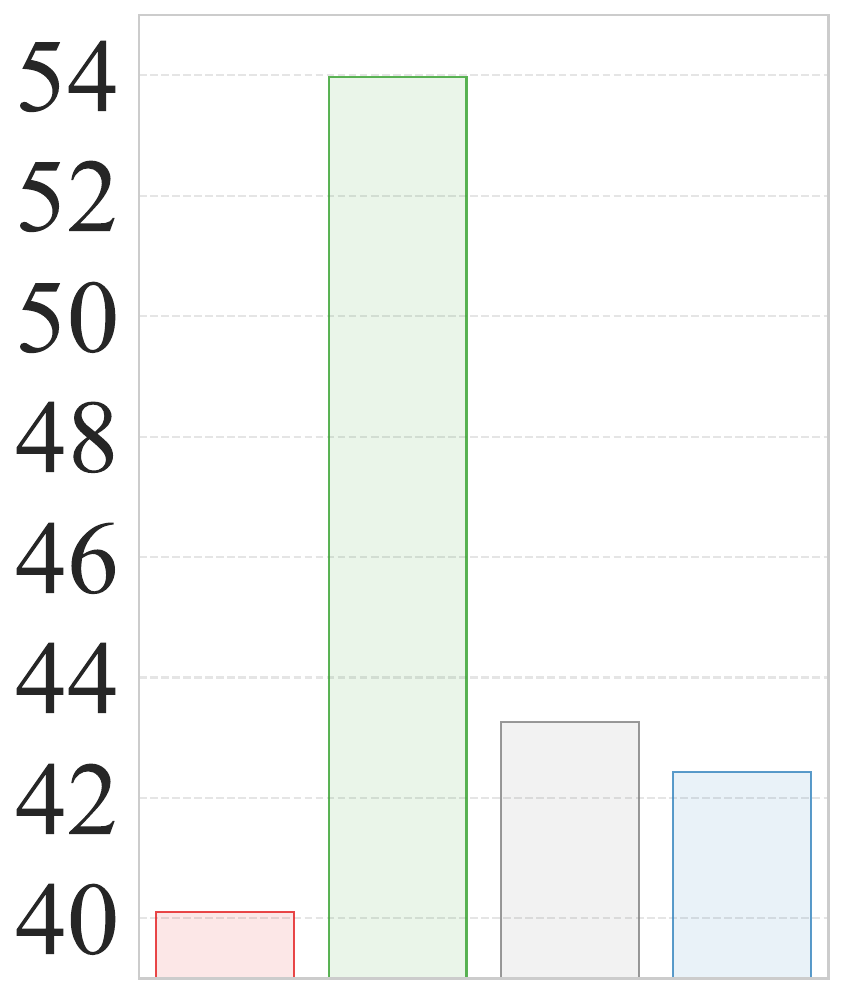}}} &
         \multicolumn{4}{c}{\multirow{5}[2]{*}{\includegraphics[scale=0.067]{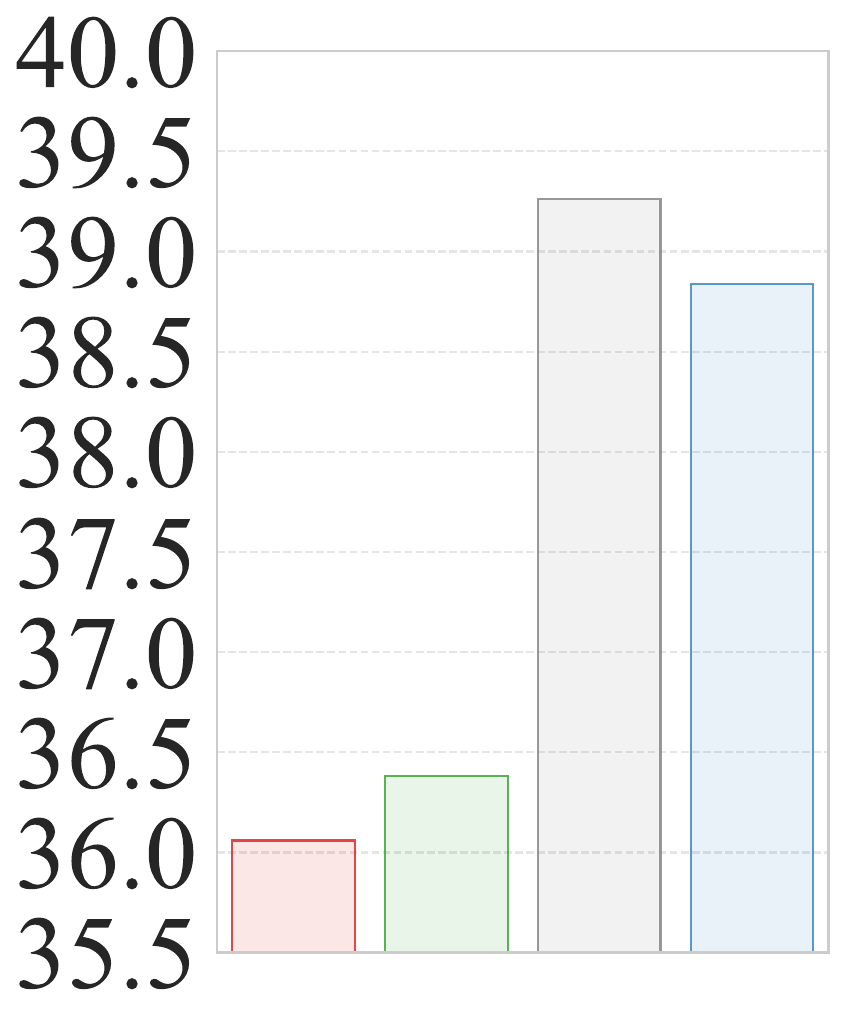}}} & 
\multicolumn{4}{c}{\multirow{5}[2]{*}{\includegraphics[scale=0.067]{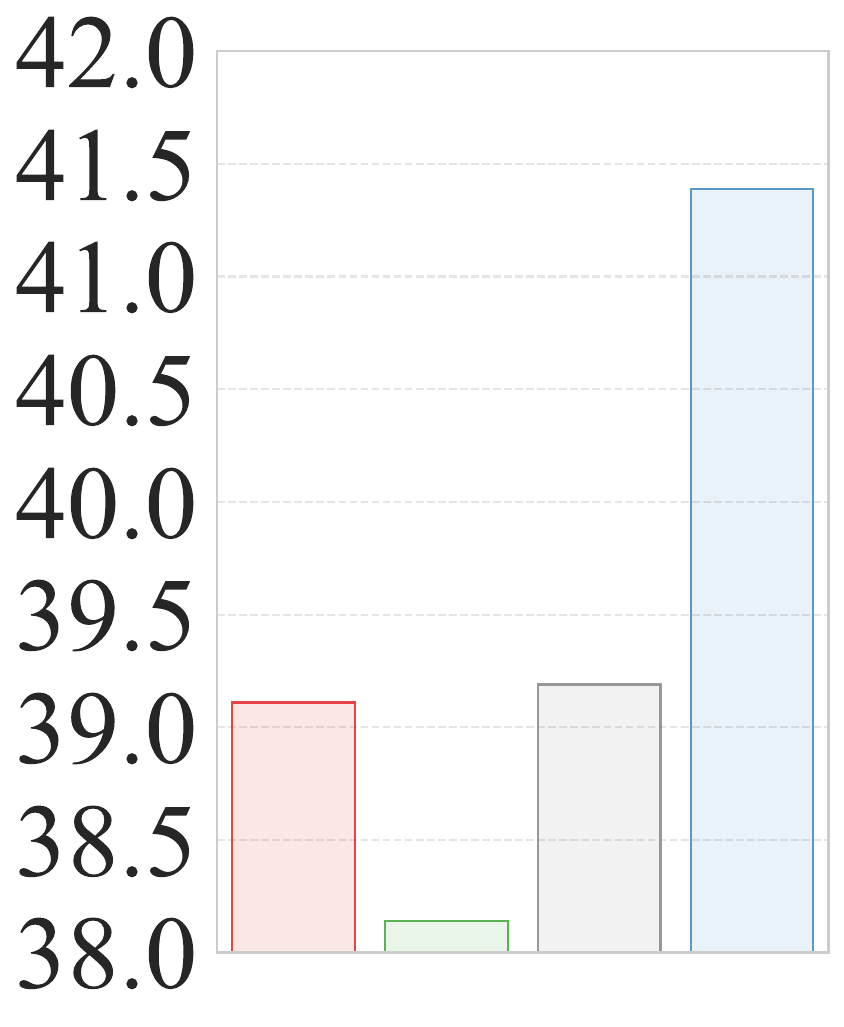}}} &
         \multicolumn{4}{c}{\multirow{5}[2]{*}{\includegraphics[scale=0.067]{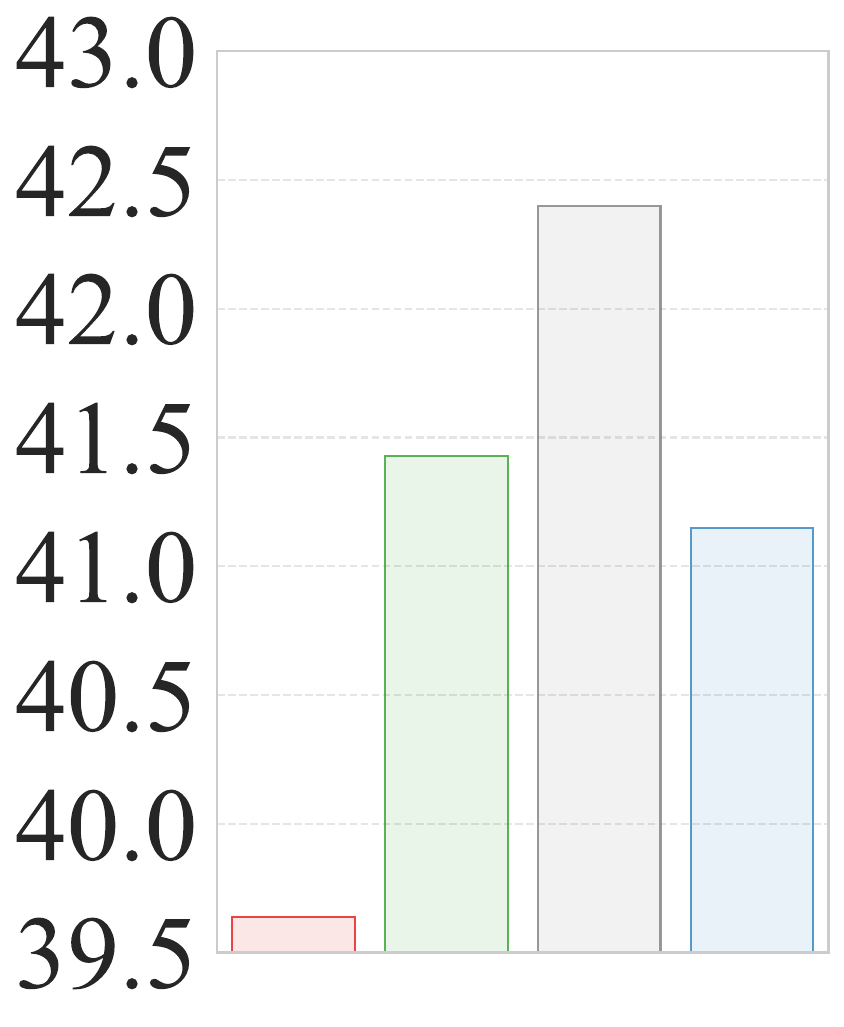}}} & 
         \multicolumn{4}{c}{\multirow{5}[2]{*}{\includegraphics[scale=0.067]{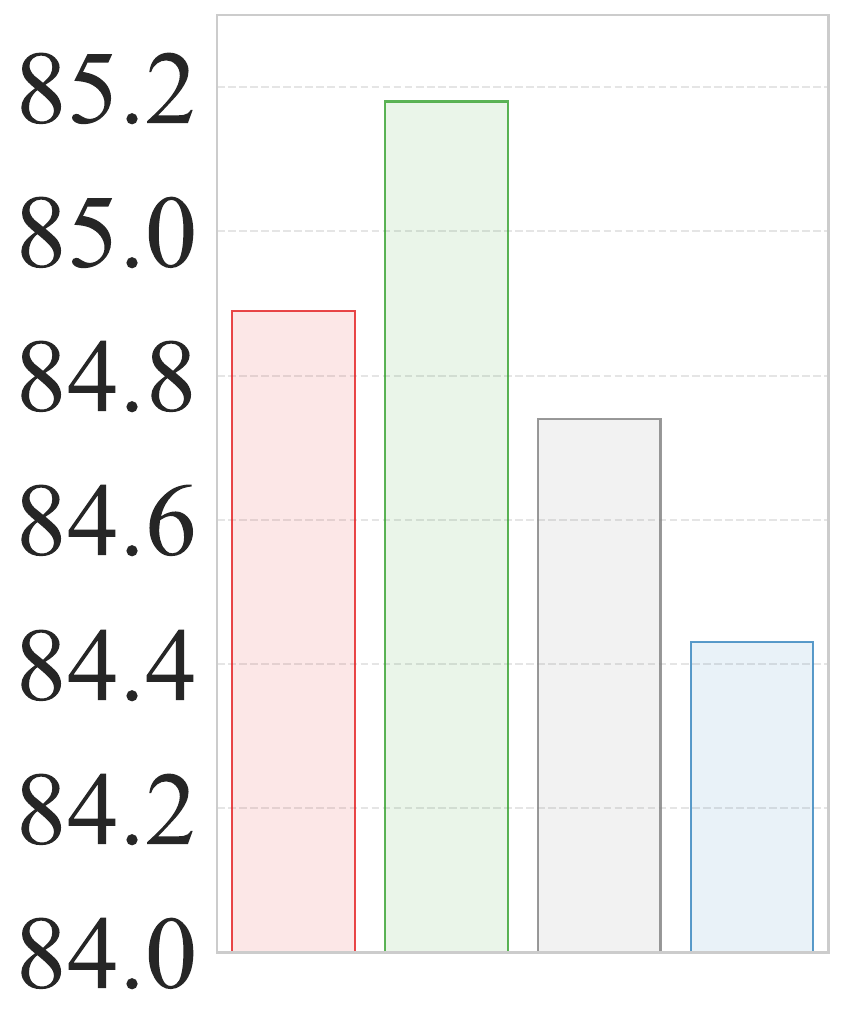}}} & 
\multicolumn{4}{c}{\multirow{5}[2]{*}{\includegraphics[scale=0.067]{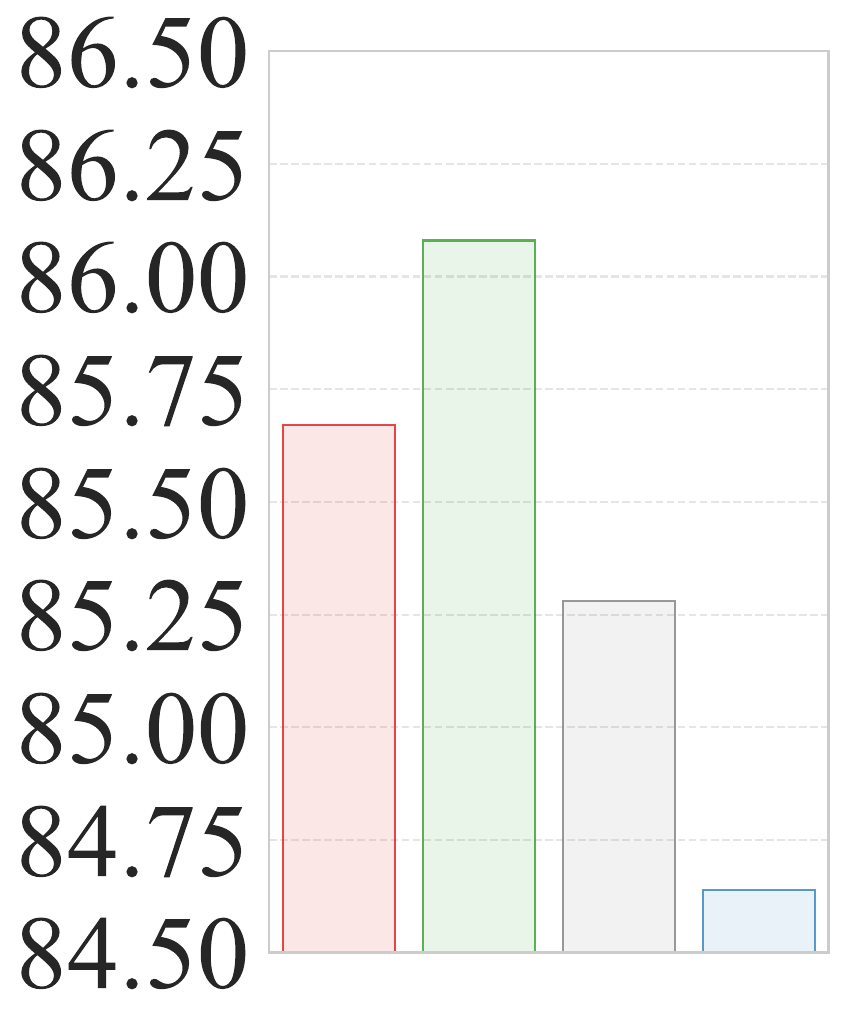}}} &
         \multicolumn{4}{c}{\multirow{5}[2]{*}{\includegraphics[scale=0.067]{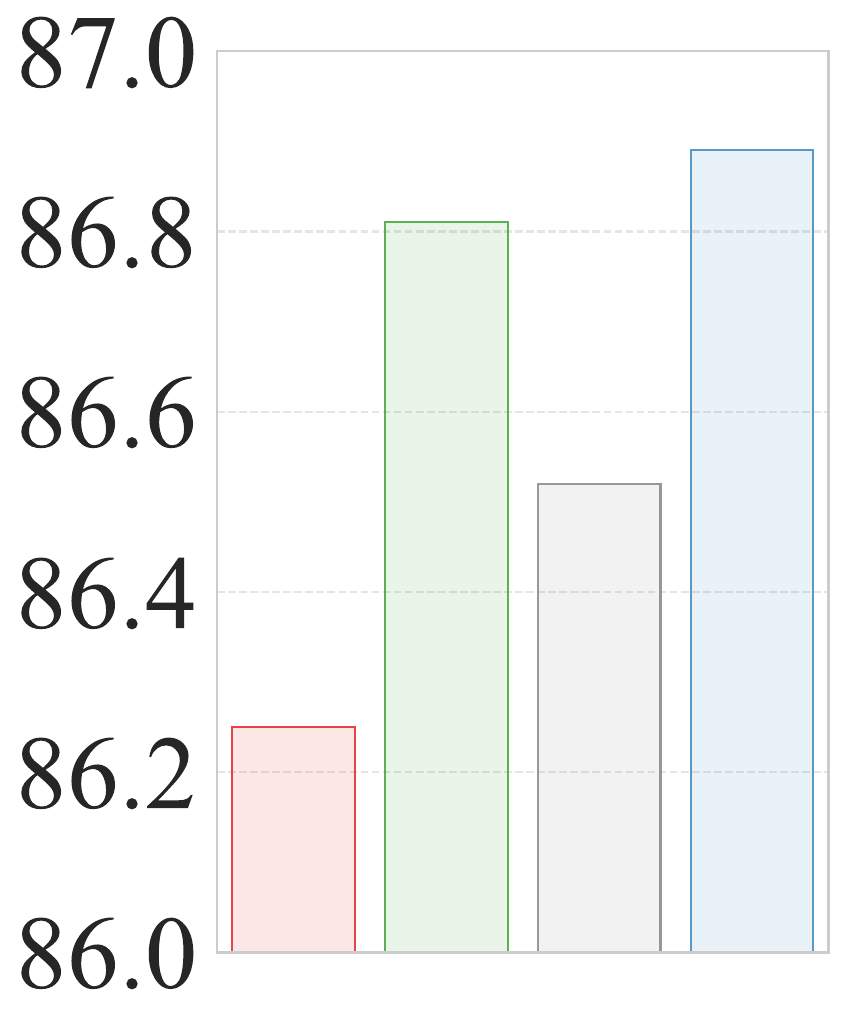}}} & 
         \multicolumn{4}{c}{\multirow{5}[2]{*}{\includegraphics[scale=0.067]{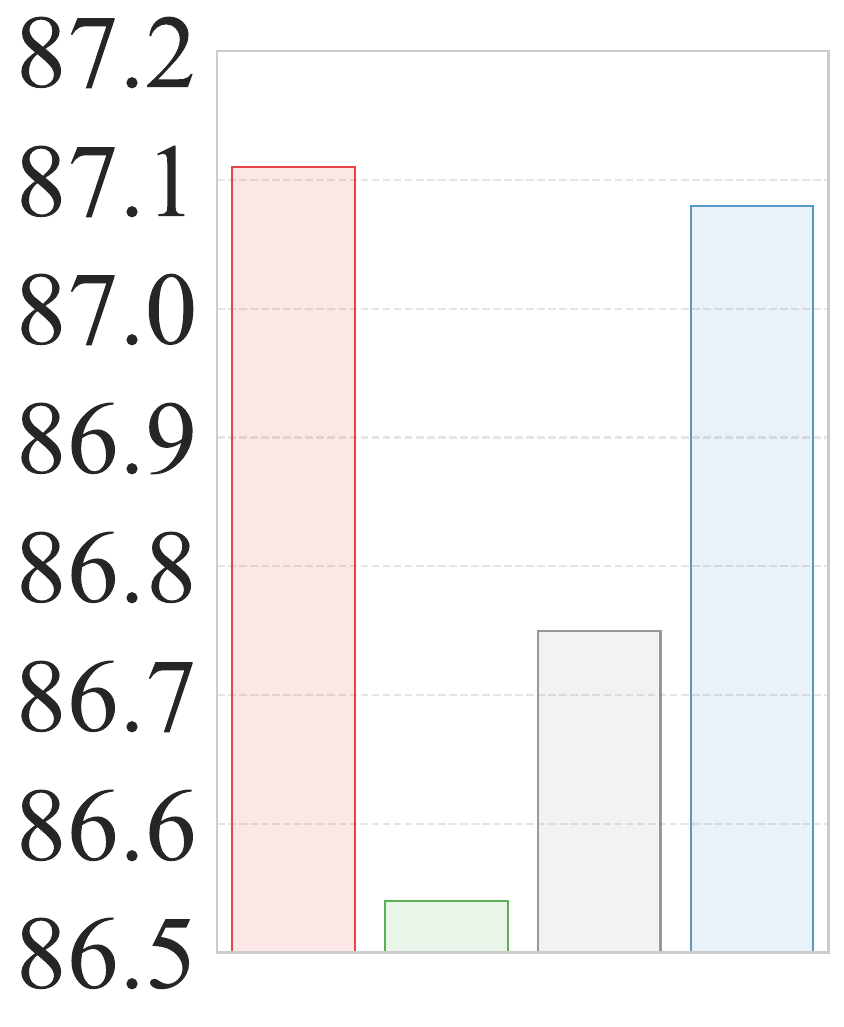}}} 
\\ \\ \\ \\ \\ \\

\midrule
\multicolumn{56}{c}{\includegraphics[scale=0.15]{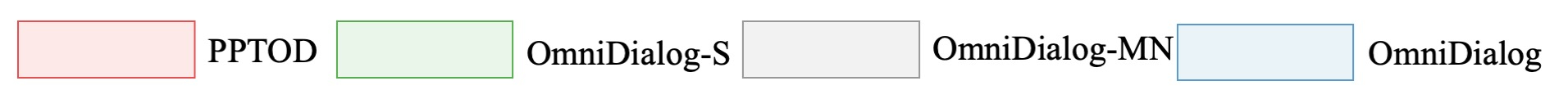}}\\
    \bottomrule 
    \end{tabular}}
    \vspace{-7pt}
  \label{tab:fine}%
\end{table*}%

\subsection{Fine-Grained Analysis}
To explore when and where the OmniDialog work, we introduce the fine-grained analysis for E2E Modeling, DST, and dialogue summarization tasks. 

\paragraph{Evaluation Aspects.}
The tasks studied in this section belong to the dialogue domain and are limited to conversations between two speakers (AI v.s. Human, or Human v.s. Human), and we call them speaker1 and speaker2, respectively.
% The tasks for fine-grained analysis are a dialogue between two speakers (AI v.s. Human, or Human v.s. Human), and we call them speaker1 and speaker2, respectively. 
% For the DialogSum, the speakers are two persons. For the E2E dialogue modeling, TODSum, and DST tasks, the two speakers are a person and an assistant.
Below, we list the aspects under exploration along with their definitions.
\begin{itemize}
    \item \textit{average speaker1 length (sp1\_len)}: the average number of words of a response for speaker1 in a dialogue. 
    \item \textit{average speaker2 length (sp2\_len)}: the average number of words of a response for speaker2 in a dialogue.
    \item \textit{average utterance length (refe\_len)}: the average number of words in the reference summary of a dialogue.
    \item \textit{average utterance number (utr\_num)}: the average number of utterances in a dialogue.
\end{itemize}

\paragraph{Settings} 
We divide the test sets of each dataset into four groups based on the sample's aspect value. Here, we explore four different models: PPTOD \cite{su2021multi}, OmniDialog-S, OmniDialog-MN, and OmniDialog, which is introduced in Sec.~\ref{sec:baselines}. All of these models employ the T5  \textsubscript{base} backbone.

\paragraph{Obervations}
The fine-grained evaluation results are shown in Tab. \ref{tab:fine}. 
The main observations are listed below:
\begin{itemize}
    \item[(1)] In general, OmniDialog demonstrates a well-balanced and enhanced ability compared to three other models, while PPTOD primarily excels in the task of end-to-end dialogue modeling and dialogue state tracking. Additionally, OmniDialog-S exhibits superior performance on two summarization datasets. By comparing it with the strong baseline model PPTOD, we ascertain that the inclusion of dialogue comprehension pre-training data significantly improves performance across all tasks.
    
    \item[(2)] OmniDialog-S and OmniDialog-MN achieve a lower combined score compared to PPTOD. By incorporating NUP and MCQA, the model gains the ability to discern whether the generated reply serves as the next utterance, indicating that additional pre-training data contribute to "teaching" the model to generate responses of higher quality.

    \item[(3)] For DST task, OmniDialog-MN exhibits relatively superior performance across various aspects. The incorporation of summarization data is believed to enhance OmniDialog-MN's effectiveness in extracting valuable information from dialogues. Furthermore, OmniDialog demonstrates a well-balanced proficiency across all aspects, showcasing its strong capability in extracting dialogue state even from lengthy conversations.

    \item[(4)] On the DialogSum dataset, OmniDialog exhibits exceptional capability, particularly in generating summaries. When it comes to long dialogues, OmniDialog-MN outperforms three other models, showcasing its advantage. By comparing the data length of DialogSum with that of the pre-training data, we observe that DialogSum has a dialogue length similar to MediaSum and longer than other pre-training datasets. We hypothesize that the MediaSum dataset contributes to teaching OmniDialog to generate longer summaries, while the inclusion of NUP and MCQA mitigates this advantage.

\end{itemize}

\section{Conclusion}
In this paper, we introduce OmniDialog, a multi-task pre-training framework that encompasses response generation, dialogue state tracking, dialogue policy learning, intent classification, multi-choice question answering, next utterance prediction, and dialogue summarization, covering dialogue management, dialogue generation, and dialogue comprehension tasks. Furthermore, we evaluate the OmniDialog on E2E dialogue modeling, DST, IC, and SUMM in various experimental settings, such as domain transfer learning, low resource scenarios, and high resource scenarios. Additionally, we conduct a fine-grained analysis to examine its performance in different aspects. The evaluations consistently demonstrate that OmniDialog exhibits comprehensive proficiency in all four selected downstream tasks.

{\appendix[Baselines]

In this section, we will make the introduction of our model much more detailed.

\paragraph{End-to-End Dialogue Modeling.} 
We compare the performance of the E2E dialogue modeling task on our model with several previous works, 
including  DAMD \cite{zhang2020task}, MinTL \cite{lin2020mintl},  DoTS \cite{jeon2021domain}, DORA \cite{jeon2022dora}, SimpleTOD \cite{hosseini2020simple}, UBAR \cite{yang2021ubar}, JOUST \cite{tseng2021transferable}, SOLOIST \cite{peng2020soloist}, TOP and TOP+Noisy Online Decoding (TOP+NOD) \cite{liu2021pretraining},  LAVA \cite{lubis2020lava}, and PPTOD \cite{su2021multi}.

DAMD \cite{zhang2020task} is an end-to-end model for multi-domain response generation, which leverages a multi-action data augmentation framework. MinTL \cite{lin2020mintl} is a transfer learning framework, that allows to jointly learn dialogue state tracking and dialogue response generation. DoTS \cite{jeon2021domain} is a task-oriented dialogue system using a simplified input context instead of the entire dialogue history, and tracks the domain state in addition to the belief state. DORA \cite{jeon2022dora} utilizes supervised learning and reinforcement learning to optimize dialogue systems with a recurrent policy. SimpleTOD \cite{hosseini2020simple} uses a single, causal language model trained on all sub-tasks recast as a single sequence prediction problem. UBAR \cite{yang2021ubar} is a dialogue system fine-tuned from GPT-2 on the sequence composed of user utterance, belief state, database result, system act, and system response of every dialog turn. JOUST \cite{tseng2021transferable}  pre-trains two agents by interacting with each other using reinforcement learning with structured reward functions on source domain dialogues, and further fine-tuning on target domain data. SOLOIST \cite{peng2020soloist} is a task-oriented bot-building paradigm based on a Transformer-based model that subsumes different dialog modules (NLU, DST, POL, and NLG) into a single model. TOP and TOP+NOD \cite{liu2021pretraining} are direct decoding for task-oriented dialogue and decoding with a noisy channel model during beam search separately. LAVA \cite{lubis2020lava} is an unsupervised approach for optimizing the latent action representation for dialogue policy optimization with reinforcement learning. PPTOD \cite{su2021multi} is a unified plug-and-play model for task-oriented dialogue with a creative dialogue multi-task pre-training strategy.

\paragraph{Dialog State Tracking.}
For dialogue state tracking, we consider the following baselines: TRADE \cite{wu2019transferable}, COMER \cite{ren2019scalable}, DSTQA \cite{zhou2019multi}, SOM-DST \cite{kim2019efficient}, MinTL \cite{lin2020mintl}, SOLOIST \cite{peng2020soloist}, UBAR \cite{yang2021ubar},  and PPTOD \cite{su2021multi}.

TRADE \cite{wu2019transferable} generates dialogue states from utterances using a copy mechanism when predicting triplets not encountered during training.  COMER \cite{ren2019scalable} treats the DST task as a sequence generation problem, instead of a pair-wise prediction problem. DSTQA \cite{zhou2019multi} models multi-domain DST as a question-answering problem. SOM-DST \cite{kim2019efficient} separates DST into two sub-tasks and guides the decoder to focus only on one of the tasks, thus reducing the burden of the decoder.

\paragraph{Intent Classification.}
 As for the intent classification task, we compare our method with several previous methods. We select BERT-Fixed \cite{casanueva2020efficient}, BERT-Tuned \cite{casanueva2020efficient}, USE+ConveRT \cite{casanueva2020efficient}, USE \cite{yang2019multilingual},  SOLOIST \cite{peng2020soloist}, TOD-BERT \cite{wu2020tod}, ConveRT \cite{henderson2019convert}, and PPTOD \cite{su2021multi}.

 BERT-Fixed \cite{casanueva2020efficient} is a BERT model with a fixed encoder, and BERT-Tuned \cite{casanueva2020efficient} is a fine-tuned BERT on IC task. USE \cite{yang2019multilingual} and ConveRT \cite{henderson2019convert} are both universal sentence encoders, tested on the IC task. USE+ConveRT \cite{casanueva2020efficient} is an intent detector based on both USE and ConveRT. TOD-BERT \cite{wu2020tod} is a pre-trained task-oriented dialogue BERT based on a contrastive objective function to simulate the response selection task.

\paragraph{Dialogue Summarization. }
For the dialogue summarization task, we test on 2 datasets. To compare on TODSum \cite{zhao2021todsum}, we select Lead-3 \cite{see2017get}, using the first three sentences of the article as a summary, BertExt \cite{liu2019text}, which is a BERT-based extractive model, Oracle \cite{narayan2018don}, using a greedy algorithm to obtain an oracle summary for each document to train extractive models, BertAbs w. DS(oracle) \cite{zhao2021todsum}, a BERT-based abstractive model with oracle dialogue state contained in input, and BART w. DS(oracle) \cite{zhao2021todsum}, a BART-based model with oracle dialogue state as the baselines. 

For DialogSum, we select LEAD \cite{see2017get}, LONGEST \cite{gliwa2019samsum},  treating the longest utterance as a summary, EXT-ORACLE \cite{narayan2018don}, Transformer \cite{vaswani2017attention}, taking Transformer as a non-pretrained abstractive baseline,  DistilBART \cite{shleifer2020pre}, which is a distilled BART model,  and PEGASUS \cite{zhang2020pegasus},  pre-trained on Transformer-based models with a new self-supervised objective.

We pre-train 2 other versions of OmniDialog: OmniDialog-S and OmniDialog-MNS. OmniDialog is trained on 7 tasks, and OmniDialog-S means "without dialogue summarization in pre-training data". Similarly, OmniDialog-MNS means "without Comprehension data, including MCQA and NUP in pre-training data".

}

\bibliography{example}
\bibliographystyle{IEEEtran}

\end{document}